\documentclass{article}
\usepackage[preprint]{neurips_2024}

\usepackage[utf8]{inputenc} 
\usepackage[T1]{fontenc}    
\usepackage{hyperref}       
\usepackage{url}            
\usepackage{booktabs}       
\usepackage{amsfonts}       
\usepackage{nicefrac}       
\usepackage{microtype}      
\usepackage{xcolor}         
\usepackage{amsmath} 
\usepackage{amsfonts}       
\usepackage{graphicx}       
\usepackage{float}
\usepackage{caption}
\usepackage{subcaption}

\title{Scaling Laws for State Dynamics in Large Language Models}

\author{%
  Jacob X Li \\
  Brown University\\
  \And
  Shreyas S Raman \\
  Brown University \\
  \AND
  Jessica Wan \\
  Brown University \\
  \And
  Fahad Samman \\
  Brown University \\
  \And
  Jazlyn Lin \\
  Brown University \\
}

\begin{document}
\maketitle

\vspace{-1em}
\section{Introduction}
\label{sec: intro}
A pivotal challenge in Reinforcement Learning (RL) \citep{maillard2013selectingstaterepresentationreinforcementlearning,Davoodi_2021,tang23g} is the emulation of an environment, which imposes \textit{constraints on state transitions} ($\mathcal{T} = Z\times Z: z_{t}\rightarrow z_{t+1}$), and a world model \citep{ha2018}, that learns to approximate true \textit{transition dynamics} ($z_{t+1} = f(z_{t}, a_{t})$) using \textit{state representations} ($z_{t} = q_{\phi}$) with sufficient information for decision making. 

SoTA world models, like PlaNet \citep{hafner2019learninglatentdynamicsplanning} and DreamerV3 \citep{hafner2024masteringdiversedomainsworld}, leverage auto-regressive architectures to model state transitions using sequential dependencies on previous states ($P(z_{t+1}|z_{t},a_{t})$). Architectures like variational autoencoders (VAEs), transformers and recurrent neural networks (RNNs) use recurrent feedback, hidden-state (memory) and attention to model local and global interactions across latent states, enabling accurate next-state predictions over long-sequences \citep{ha2018, Schrittwieser_2020}.  Accurate world models have shown to improve efficiency \citep{hafner2019learninglatentdynamicsplanning} in long-horizon decision making and context-awareness in state representations.

Large Language Models (LLMs) \citep{brown2020languagemodelsfewshotlearners,touvron2023llamaopenefficientfoundation,deepseekai2025deepseekr1incentivizingreasoningcapability} can be viewed as a `world model' for text that learn latent \textit{token representations} and \textit{autoregressively generate} future tokens ($P(t_{i}|t_{i-1},...,t_{1})$) conditioned on previous tokens. Furthermore, residual streams in Transformers preserve information across layers with linear updates reflecting interpretable/traceable modifications to an `internal state' \citep{vaswani2023attentionneed,Radford2019LanguageMA}. Open questions in this space remain, such as the scale at which LLMs fail to accurately model transition-dynamics, explaining mechanisms enables success and discovering whether substructures modeling state-dynamics exist. Our study could clarify the limitations of residual-stream for state representation i.e. where it \textit{can't} be applied, or uncover powerful state-dynamics systems that do not require additional pretraining or leverage substructures for efficient decision making. 

Accordingly, we propose the following research questions: (1) What are the limits to using the residual stream to capture state-dynamics as the number of states and number of transition constraints scale? (2) How can mechanistic interpretability provide causal explanations for why predicting state-dynamics fails at certain scales? (3) What mechanisms or circuits in LLMs particularly model sequential state-dynamics?






\section{Related Works}
\subsection{State Learning and Transition Tracking in Large Language Models}
\label{subsec:state-tracking}
LLMs have demonstrated implicit abilities to track and simulate state information beyond`pattern-matching' in reasoning and planning tasks, despite lacking explicit training for these functions \citep{hao2023reasoning,zhao2023explicit,valmeekam2023planning,guo2023codeplan,yao2023react}. Internet-scale unsupervised pre-training on sequential data can yield rich state representations, observable both in synthetic environments (e.g., board games) \citep{li2023emergent,wattenberg2023gradient,sunthe2023linear} and in real-world planning scenarios \citep{rozanov2024stateact,yao2023react}. Mechanistic interpretability studies have begun to identify the circuits, neurons, and attention heads responsible for this tracking \citep{li2023emergent,kim2023entitytrackinglanguagemodels,li2025howlanguagemodelstrack}, and further work explores how these mechanisms emerge during pre-training or are enhanced via fine-tuning \citep{prakash2024finetuningenhancesexistingmechanisms}. Nevertheless, recent benchmarks reveal persistent failures on more complex planning problems \citep{zuo2025planetariumrigorousbenchmarktranslating,silver2023generalizedplanningpddldomains,liu2023llmpempoweringlargelanguage}, motivating our investigation into the roots of these deficiencies.

\subsection{LLM Capabilities and Scaling Effects}
\label{subsec:llm-scaling}
Empirical scaling laws show that increasing model size and training data volume yields consistent performance improvements and emergent abilities beyond critical parameter thresholds \citep{wei2022emergent, schaeffer2023abilities}. However, the relationship between model scale and the ability to track discrete state dynamics remains largely unexplored \citep{georgetown2024emergent, hao2023reasoning}.

\subsection{Decision Making Using LLMs and Autoregressive Models}
\label{subsec:decision-making-llm}
Autoregressive LLMs have been shown to adhere to formal grammars in symbolic decision‐making domains (e.g.\ PDDL, LTL, text games) and understand constrains in the real-world environments, but they often lack robust understanding of environment‐specific dynamics \citep{ahn2022icanisay, tsai2023largelanguagemodelsplay}. In parallel, latent next‐state predictors such as Decision Transformer and PlaNet improve sample efficiency in model‐based RL \citep{chen2021decisiontransformerreinforcementlearning, hafner2019learninglatentdynamicsplanning}, yet their internal transition‐dynamics mechanisms are not very interpretable.
 
\section{Methodology}
Our goal is to systematically study how transformer-based LLMs represent and update internal state towards the core research questions proposed in Section \ref{sec: intro}. To this end, we consider $3$  domains—Box Tracking, Abstract DFA Sequences, and Complex Text Games—that can each be formally modelled as a deterministic finite automaton (DFA)\citep{RM-704}. A DFA is defined as a 5-element tuple $M = (Q, \Sigma, \delta, q_0, F)$, where $Q$ is a finite set of states, $\Sigma$ is the input alphabet, $\delta: Q \times \Sigma \rightarrow Q$ is the deterministic transition function, $q_0 \in Q$ is the initial state, and $F \subseteq Q$ denotes the set of terminal states.  

We run $3$ experiments on each domain across different combinations of\textit{states} $Q$ and \textit{state-action transitions} $T:=q_0 \xrightarrow{\delta(q_{0})} q_1 \xrightarrow{\delta(q_{1})} q_2 \cdots \xrightarrow{\delta(q_{t-1})} q_t$ sizes to understand LLM's state-tracking capabilities for transition-dynamics and transition-history at scale. 

\subsection{Task Domains}
\label{subsec:task-domains}
To probe different facets of state-tracking, we design three complementary tasks:

\begin{enumerate}
  \item \textbf{Box Tracking:} \label{sec:task-box}
The model is given an initial configuration such as
“The hat is in Box A. The glove is in Box B. The ball is in Box A.”
It then processes moves such as “Move the hat from Box A to Box B. Move the ball from Box A to Box B. The glove is in the Box” and must complete with the final location “B”. This task evaluates dynamic state updates with simple move semantics.
  \item \textbf{Abstract DFA:} In this domain, the model is given all valid states (lowercase letters) and actions (uppercase letters) from a randomly generated deterministic finite automaton (DFA). The model then observes a randomly sampled sequence of state-action transitions and must predict the state following the last action. Unlike other domains, the semantics of transitions are explicitly defined but transition dynamics (which are highly restricted) need to be inferred. The LLM must still accurately track a large combinatorial state space. \textbf{Prompt:} "Start at state a. Take action M, go to state b. Take action K, go to state a. Take action M, go to state" \textbf{Answer:} " b".
  \item \textbf{Complex Text Games:} In this logic-driven domain, the LLM is tasked with tracking allocations of $n$ fruits to $n$ individuals based on specified fruit assignments and transfers between people. Similar to Einstein’s Puzzles, this domain features a large combinatorial state space and requires the LLM to infer valid transitions from context, often narrowing down to a single feasible fruit assignment. Solving the task demands tracking multi-variable state updates, performing allocation and counting, and enforcing constraints inferred implicitly from the prompt. \textbf{Prompt:} "Kate, Sarah, Jack, Dean walk into a fruit store. There are only 4 fruits: grape, apple, peach, pear. Each person gets a different fruit. Sarah gives Jack the peach. Sarah can have the" \textbf{Answer:} "grape, apple, pear".
\end{enumerate}

\subsection{Experiments \& Metrics}
We run $3$ experiments respectively over varying numbers of states $|Q|$ and transitions $|T|$ -- keeping transition density ($\frac{|\delta|}{|Q|}$) fixed:
\begin{enumerate}
    \item \textbf{Representational capacity} of LLM residual streams for state-dynamics. We measure this via accuracy of next-state prediction against ground-truth next-state from the DFA i.e. comparing   $LLM(T):= \tilde{q}^{'}$ against $\delta(q) := q^{'}$. This captures the scale at which LLM's internal representation fails to track transition-dynamics or transition-history
    \item \textbf{Attribution of state-tracking} behavior of LLMs. We utilize activation patching (using logit difference) on models that have reasonable competence on each task -- from previous evaluation. This would help us understand which model components (layer or attention heads) and tokens are responsible for state-tracking behavior. 
    \item \textbf{Mechanisms for state tracking} in LLMs. We analyze the attention patterns of relevant heads (found in previous evaluation) by aggregating their attention probabilities from final token position. This would help us investigate the presence of \textit{state-dynamics heads} that move state-history or state-action dependence information across residual streams.   
\end{enumerate}
The attention heads we discover might function as internal symbolic operators amenable to probing or targeted intervention, though we defer  such techniques of intervention and probing to future work. Further details on our experiment techniques can be found in Appendix \ref{subsec:explained_experiments}.

\subsection{Models}
We evaluate $3$ autoregressive Transformer families at different parameter scales:
(1) \textbf{TinyStories} (8 M, 28 M, 33 M params) as a minimal‐capacity floor. (2) \textbf{GPT-2} (117 M, 345 M, 774 M, 1.5 B) as a canonical mid-scale reference. (3) \textbf{Pythia} (14 M, 70 M, 1 B) to fill capacity gaps.

\section{Results}
\subsection{Box Tracking Task}

\begin{figure}[!ht]
    \centering
    \includegraphics[width=0.24\linewidth]{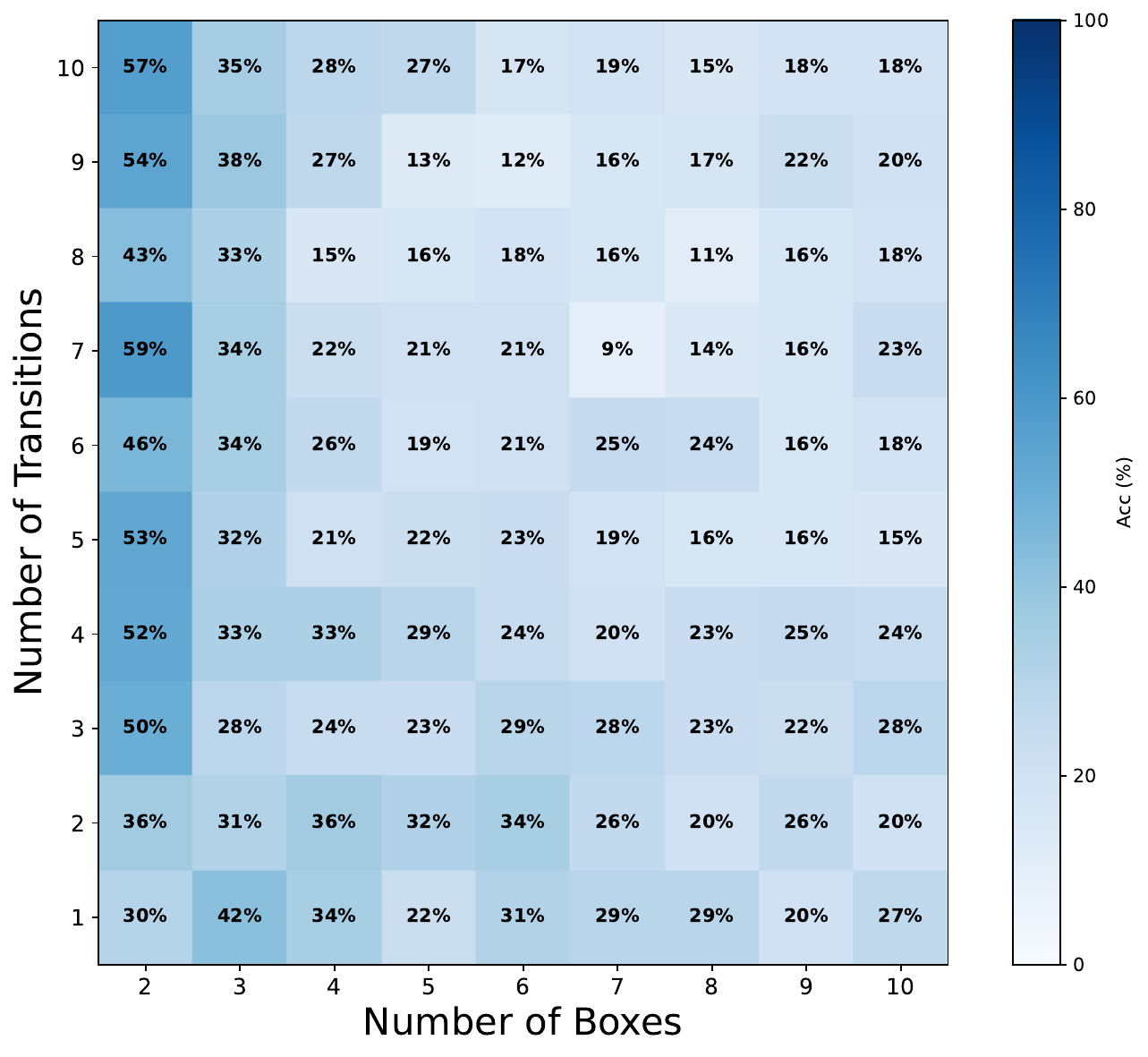}
    \includegraphics[width=0.24\linewidth]{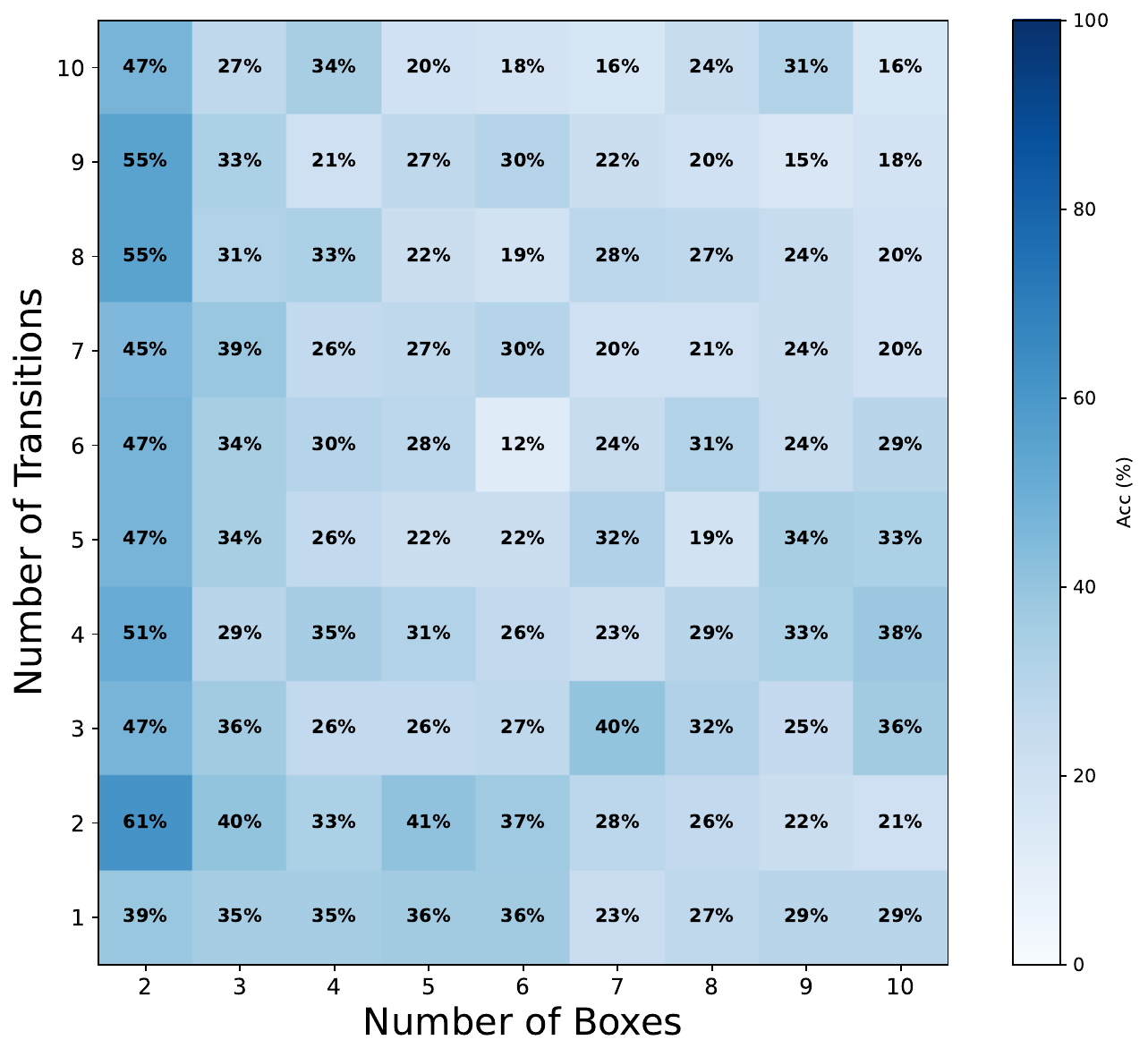}
    \includegraphics[width=0.24\linewidth]{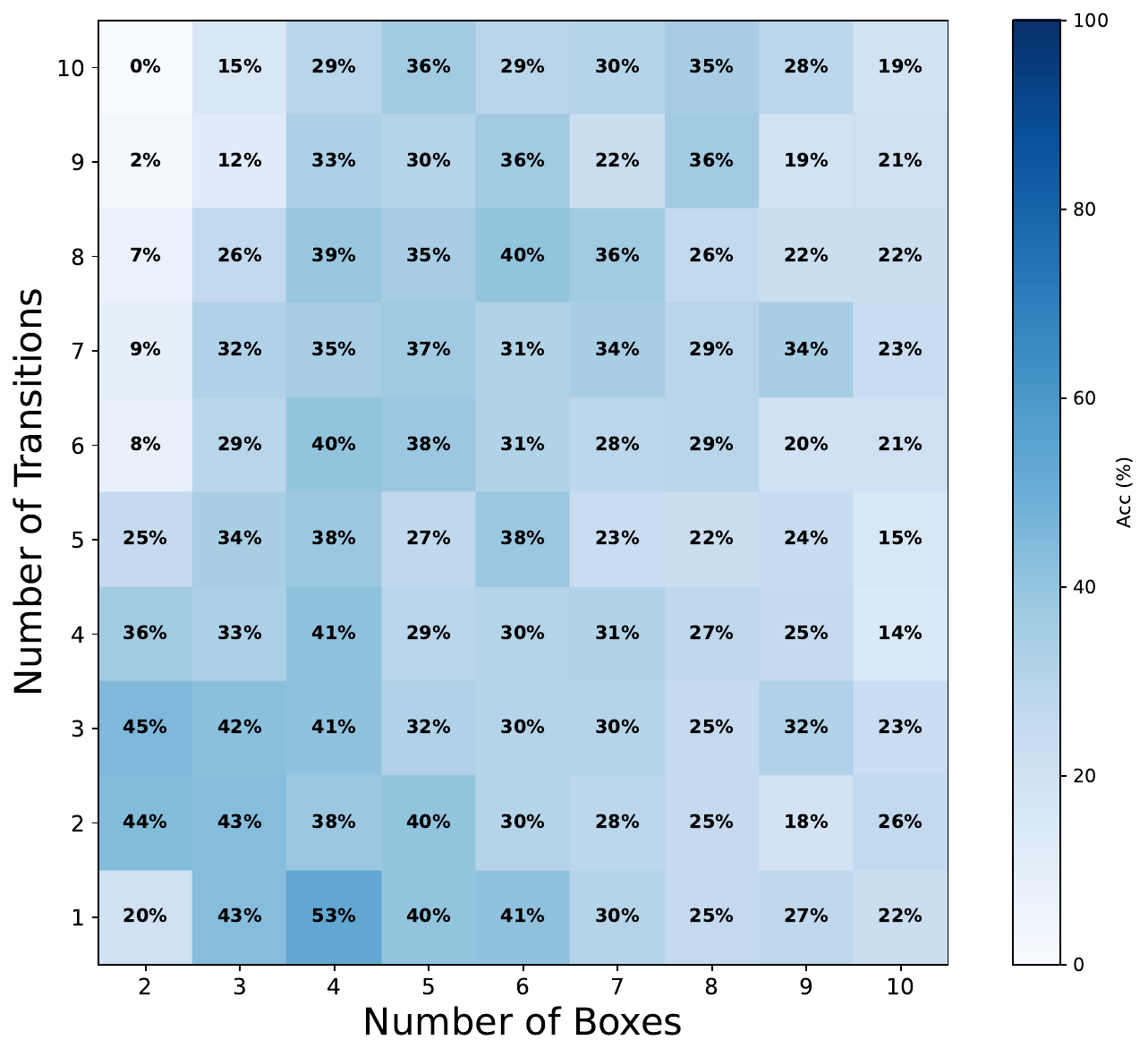}
    \includegraphics[width=0.24\linewidth]{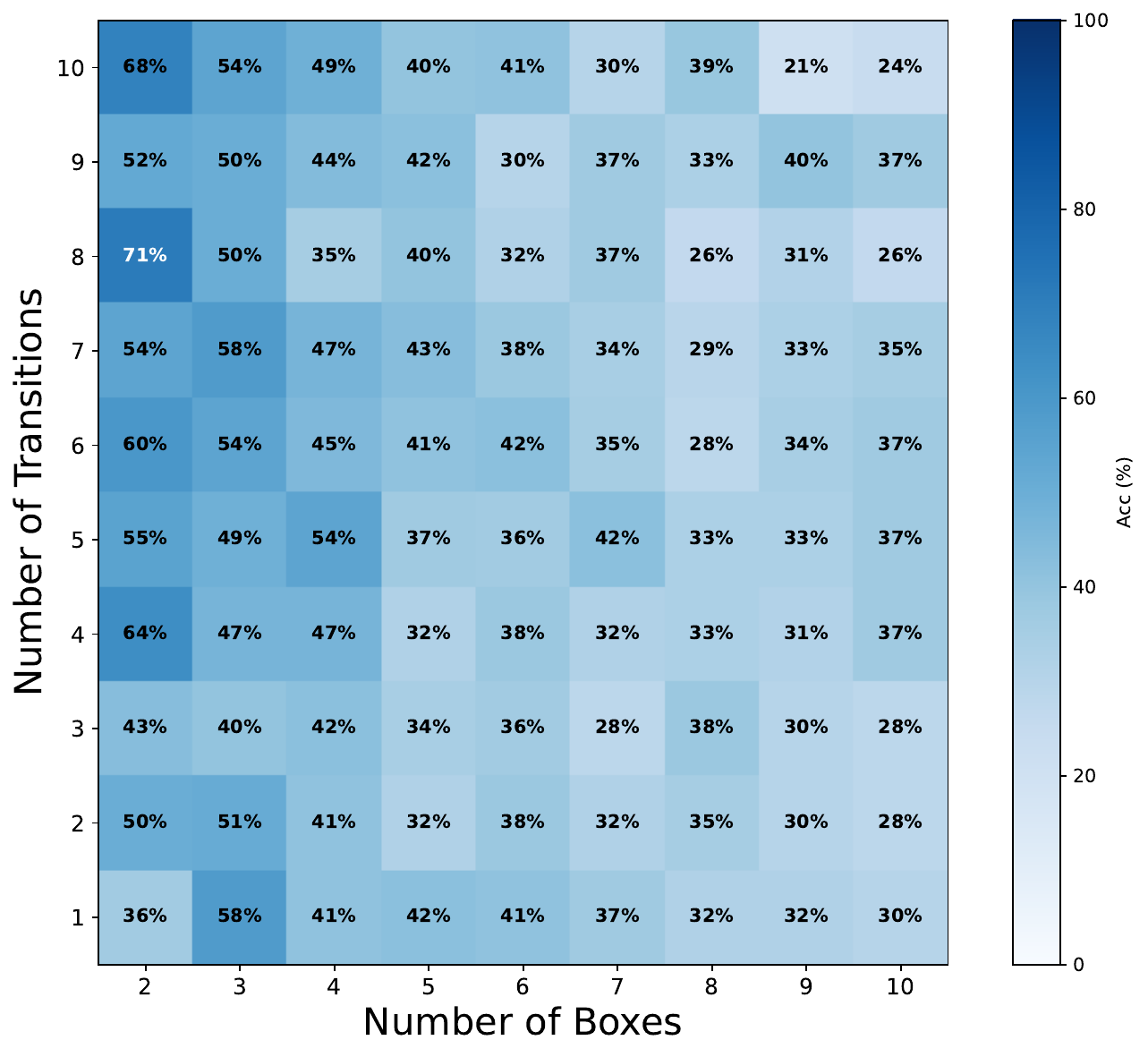}
    \caption{GPT2 Model Performance (Small, Medium, Large, X-Large)}
    \label{fig:box-GPT2}
\end{figure}

Our findings indicate a clear relationship between model scale and task performance. Larger models within a family generally exhibit better accuracy on the Box Tracking task, suggesting that increased model capacity aids in tracking and updating entity states. For instance, in Figure \ref{fig:box-GPT2}, comparing models within the GPT-2 suite, we observe an improvement in accuracy as model size increases. While larger models perform better, the overall accuracy still presents opportunities for future improvements, particularly as the number of boxes and transitions increases, thereby escalating task complexity. The detailed evaluations for the TinyStories and Pythia model families are provided in Appendix \ref{sec:ts-box} and \ref{sec:py-box}.

\subsubsection{Activation Patching to Identify Key Model Components}
To understand which model components are responsible for successfully performing the Box Tracking task, we employ activation patching. Given its relatively strong performance and manageable size for mechanistic interpretability, we selected the GPT-2 XL model for this analysis.

Using the experimental approach described in \ref{subsec:explained_experiments} , we compute and average this patching metric over a dataset of 100 such clean/corrupted pairs -- focusing on scenarios with 3 boxes, 3 objects, and 2 transitions.

\begin{figure}[!ht]
    \centering
    \begin{subfigure}[t]{0.49\textwidth}
        \centering
        \includegraphics[height=4.5cm]{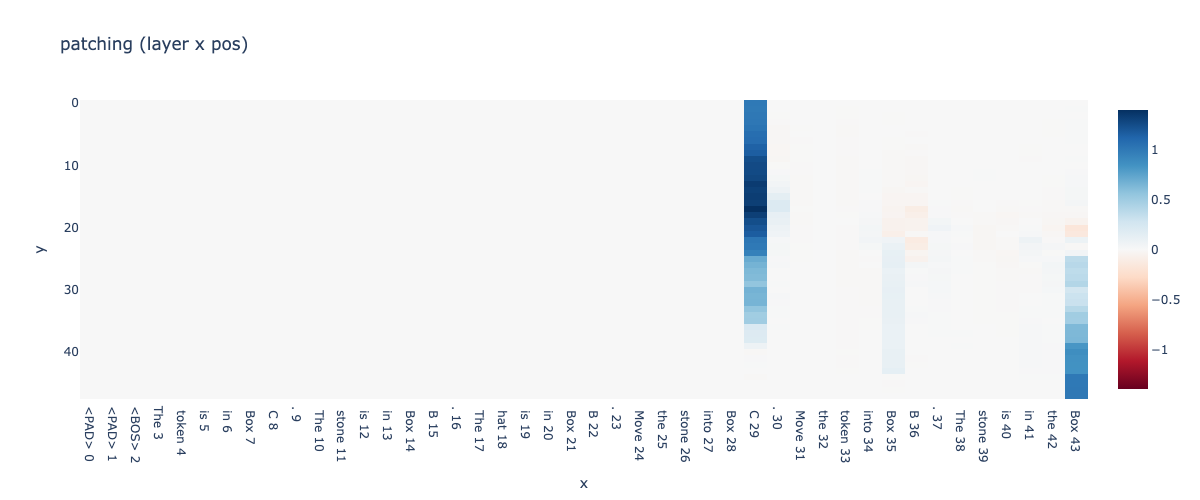}
        \caption{Activation patching of residual stream with GPT-2 XL}
        \label{fig:patching_layer_pos_heatmap_gptxl}
    \end{subfigure}%
    \hfill
    \begin{subfigure}[t]{0.49\textwidth}
        \centering
        \includegraphics[height=4.5cm]{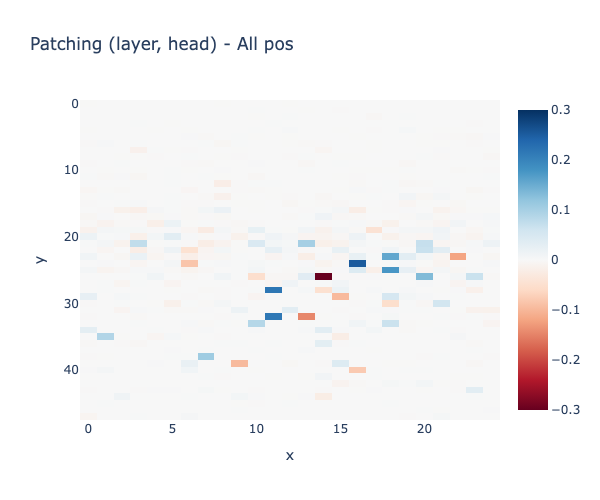}
        \caption{Activation patching for GPT-2 XL on the Box Tracking task}
        \label{fig:patching_heatmap_gptxl}
    \end{subfigure}
    \caption{Comparison of activation patching results for GPT-2 XL. Left: residual stream patching across token positions. Right: attention head patching.}
    \label{fig:combined_patching_heatmaps}
\end{figure}

Figure \ref{fig:patching_layer_pos_heatmap_gptxl} illustrates the impact of patching the residual stream at each layer (y-axis) and token position (x-axis). A key observation from this plot is the effect of patching at the token positions corresponding to the last occurrence of the queried object's name or its destination box in a move statement. The influence of patching at these critical token positions on the final answer begins to diminish around layers 20-40. This suggests that the crucial information regarding the object's state, particularly its final location, has been effectively processed and propagated to later parts of the model by this layer range. This finding aligns with the subsequent head-specific patching results.

Figure \ref{fig:patching_heatmap_gptxl} displays the results of patching the output of each attention head across all layers and token positions in GPT-2 XL. The heatmap reveals that a specific set of attention heads in the later layers are crucial for task performance.

\begin{figure}[h!]
\centering
\includegraphics[width=\textwidth]{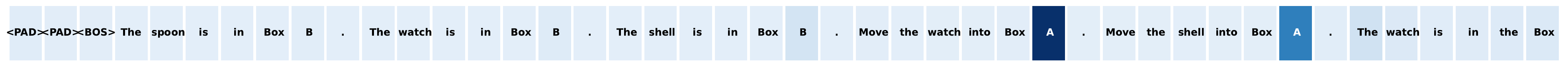} 
\caption{Aggregated attention pattern from the final token position for the top 5 heads identified by activation patching in GPT-2 XL, on an example prompt in Box Tracking domain. The queried object is "watch", and its correct final location is "Box A". The attention pattern shows strong focus on "A" especially in the phrase "Move the watch into Box A".}
\label{fig:attention_viz_gptxl}
\end{figure}

To further investigate the function of these important heads, we analyzed their attention patterns. We aggregated the attention probabilities from the final token position (the token immediately preceding the answer token "Box") of the top 5 most impactful heads identified through patching. Figure \ref{fig:attention_viz_gptxl} shows an example of such an aggregated attention pattern. The visualization highlights that these critical heads strongly attend to tokens related to the queried object's relevant initial placement and, crucially, its final destination box mentioned in the last relevant "Move" instruction. For instance, if querying the location of the "watch" which was last moved into "Box A", these heads attend to the "A" token within that "Move the watch into Box A" statement. This indicates that these heads play a significant role in identifying and propagating the final location of the queried entity.

\subsection{Abstract DFA Task}
To assess LLM's performance on state-tracking with highly-restricted transition dynamics, evaluate sequences from the Abstract DFA task using TinyStories, GPT2, and Pythia models of varying sizes. Following Section \ref{subsec:task-domains}, we systematically vary the number of states and randomly-generate transitions through the DFA to control task complexity, and report average next-state prediction accuracy below

\begin{figure}[!ht]
    \centering
    \includegraphics[width=0.25\linewidth]{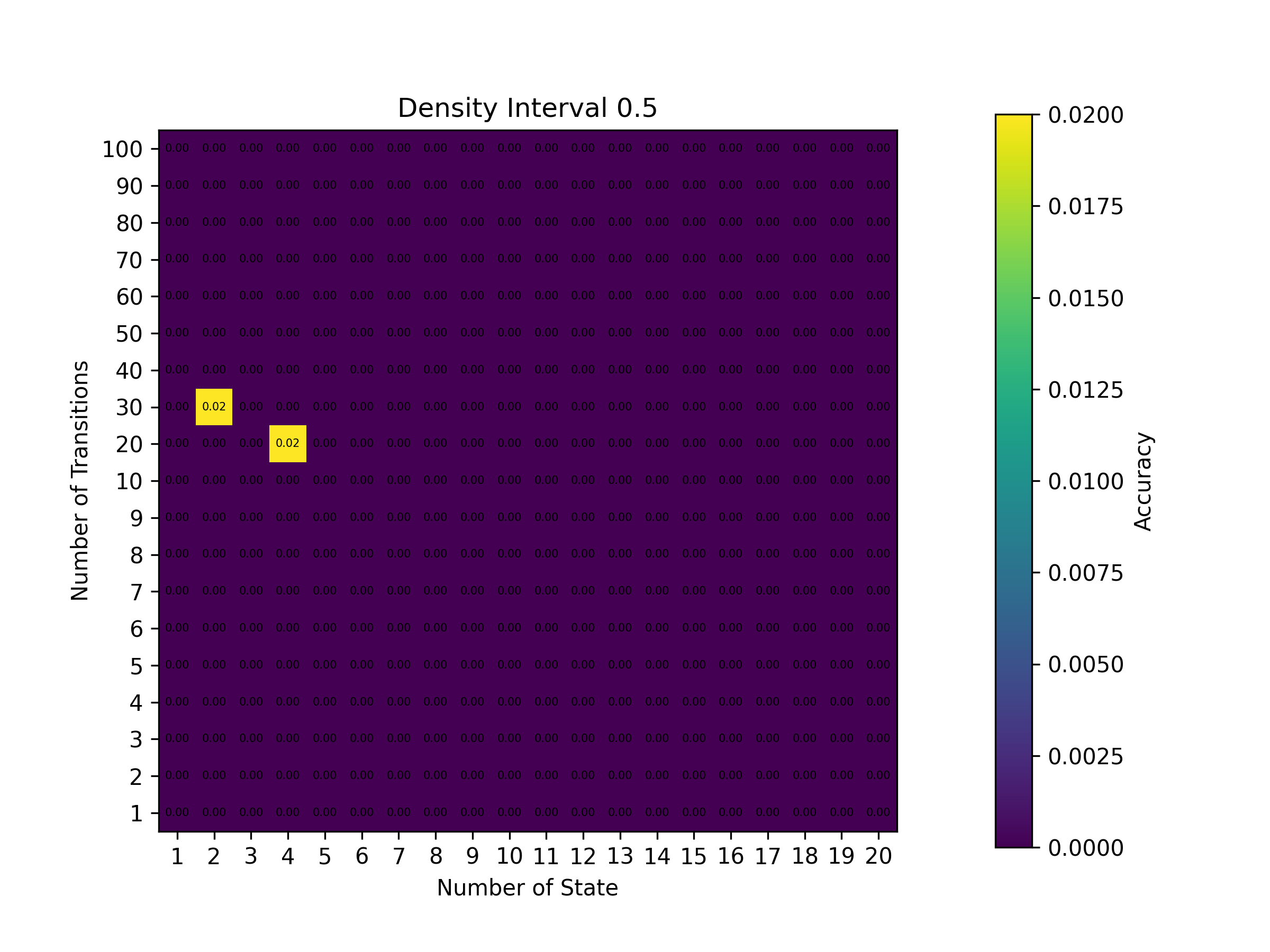}
    \includegraphics[width=0.25\linewidth]{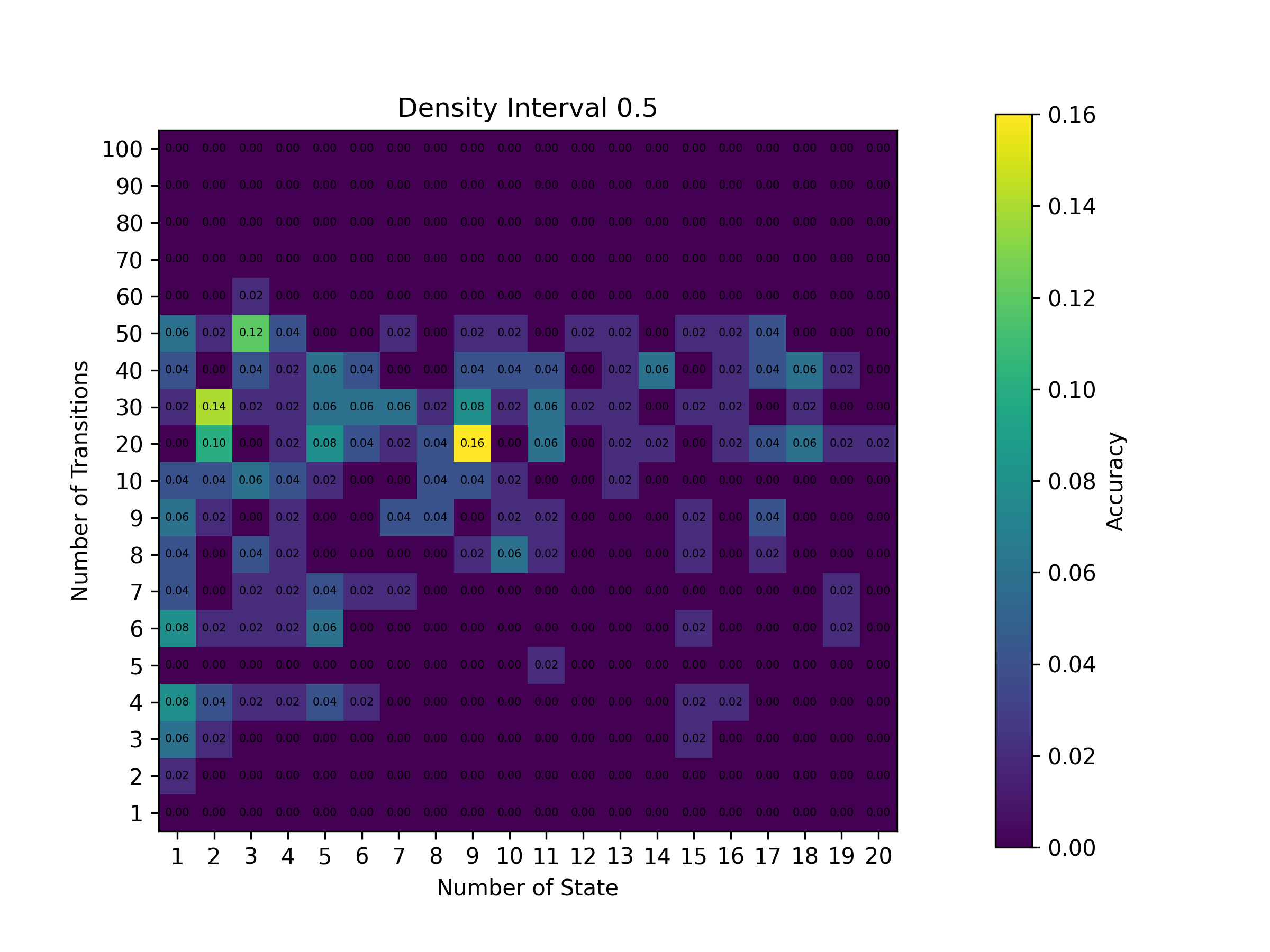}
    \includegraphics[width=0.25\linewidth]{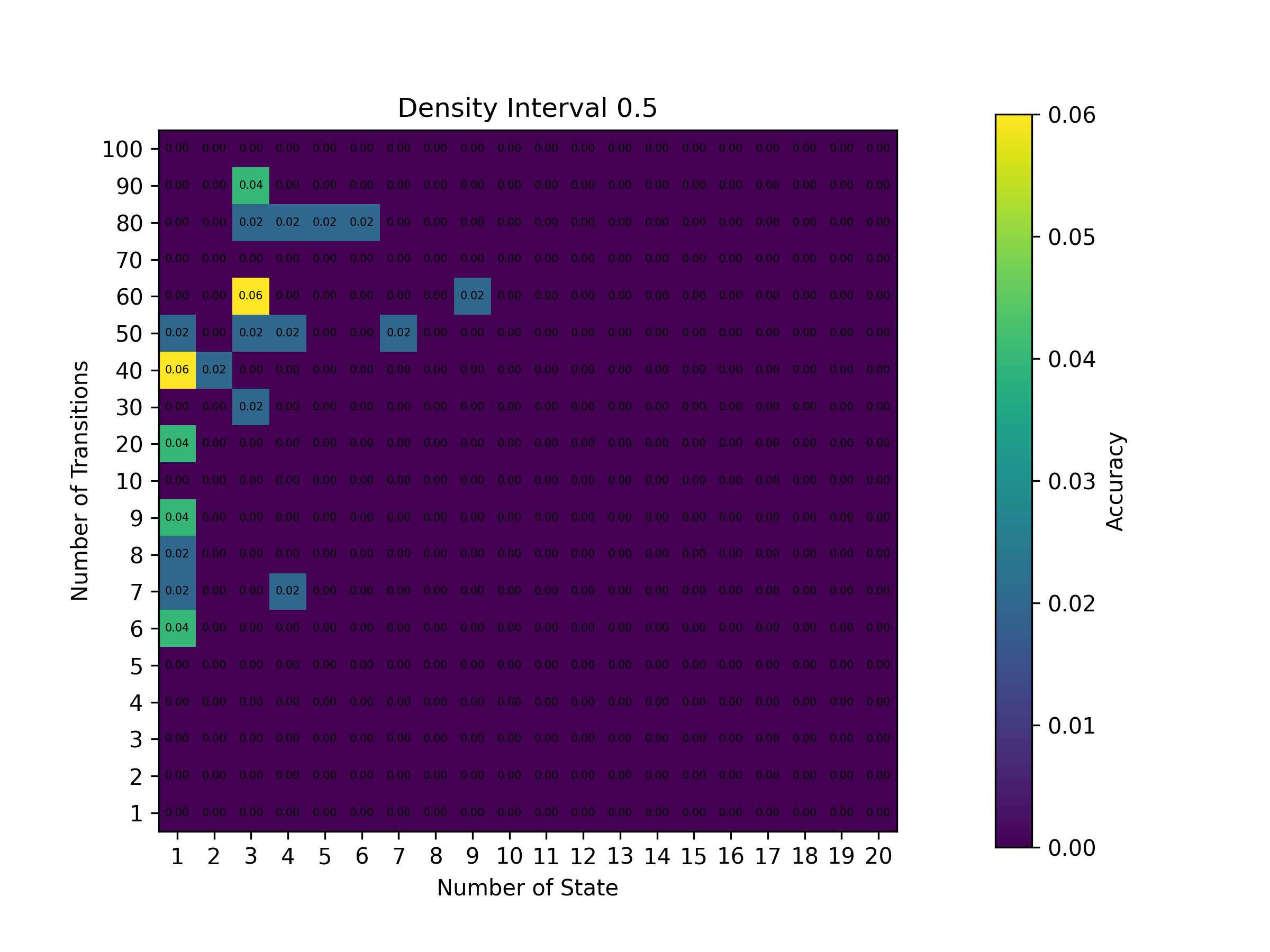}
    \caption{TinyStories Model Performance (8M, 28M, 33M)}
    \label{fig:dfa-TinyStories}
\end{figure}
\begin{figure}[H]
    \centering
    \includegraphics[width=0.24\linewidth]{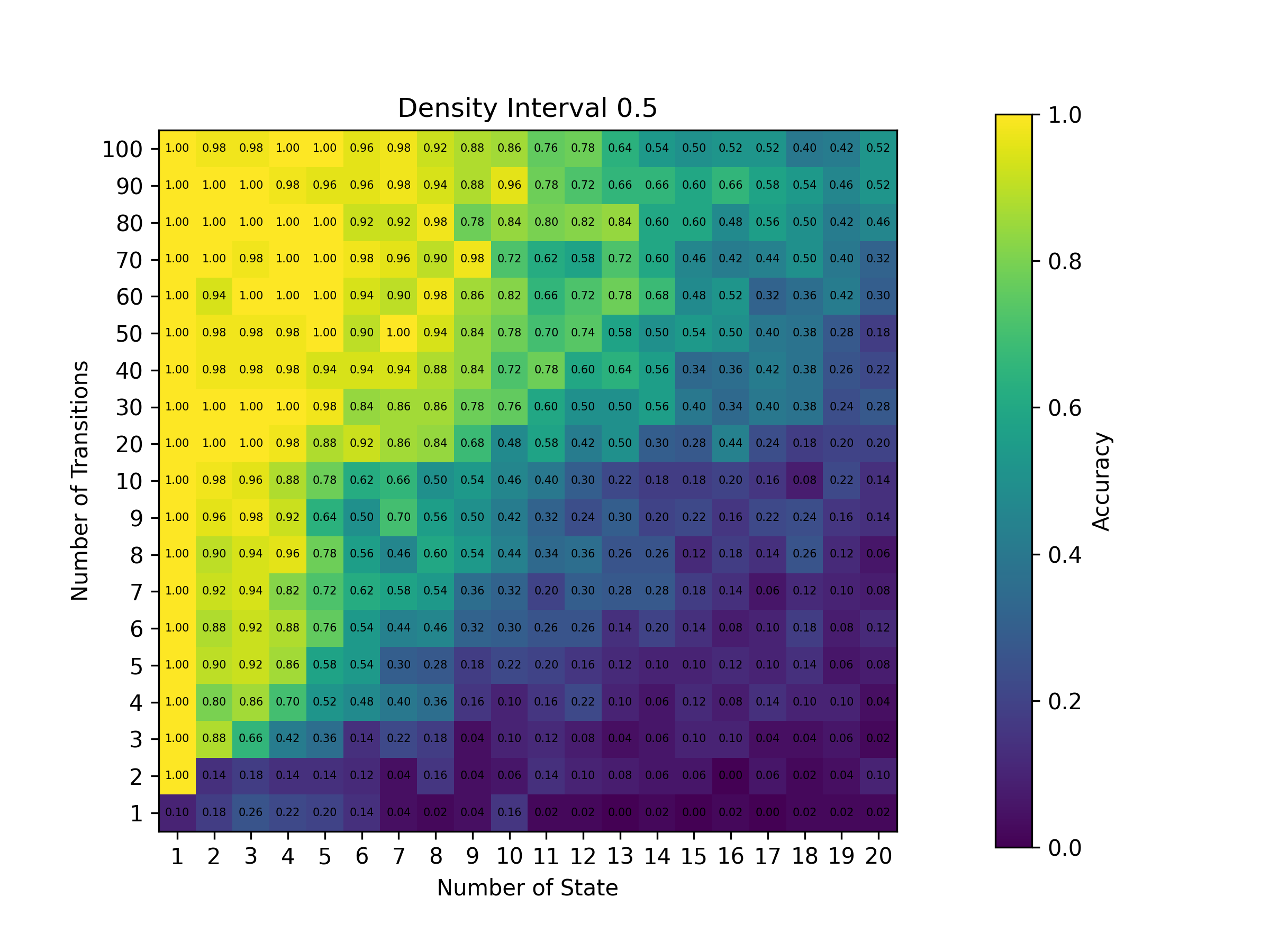}
    \includegraphics[width=0.24\linewidth]{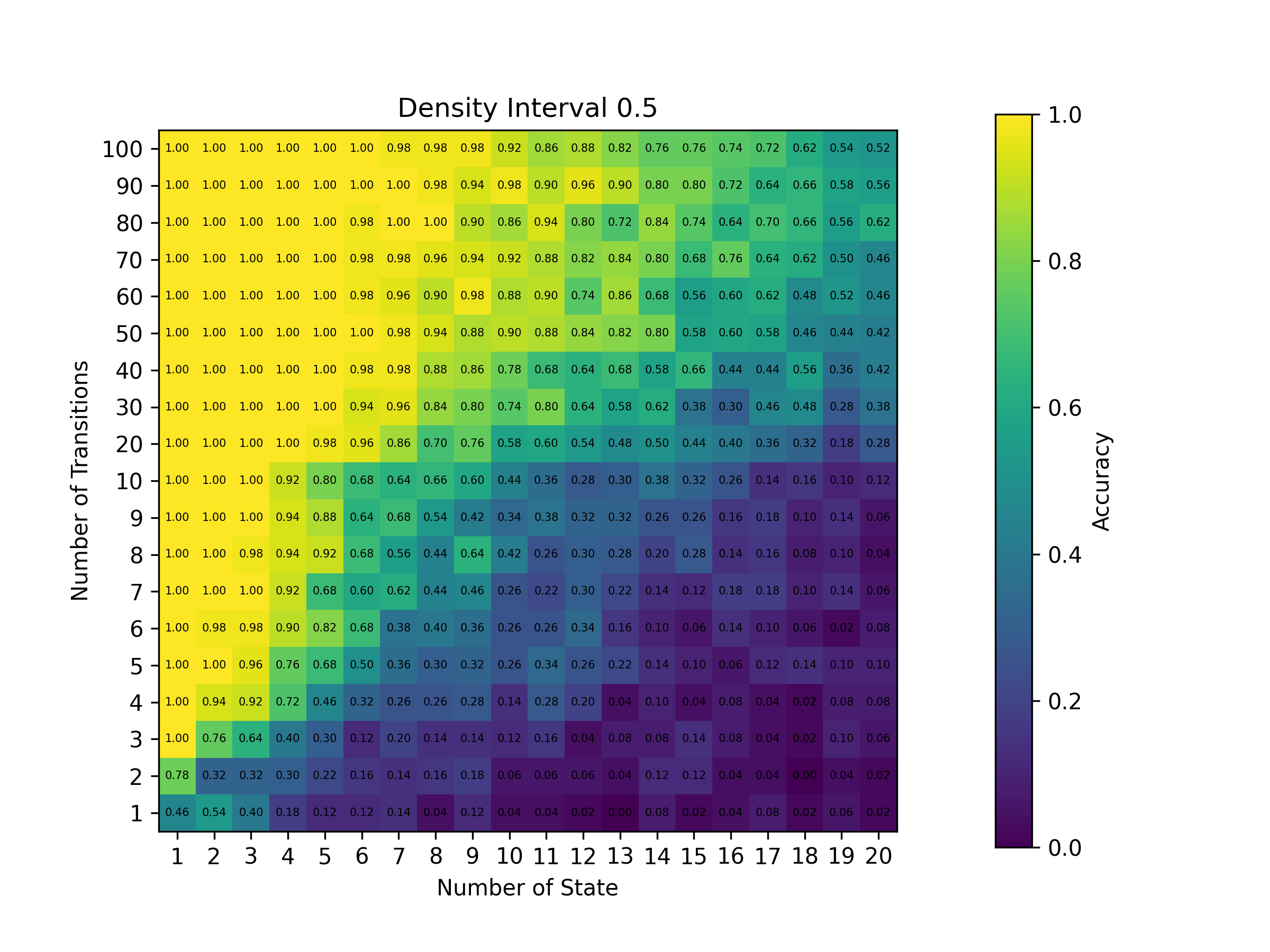}
    \includegraphics[width=0.24\linewidth]{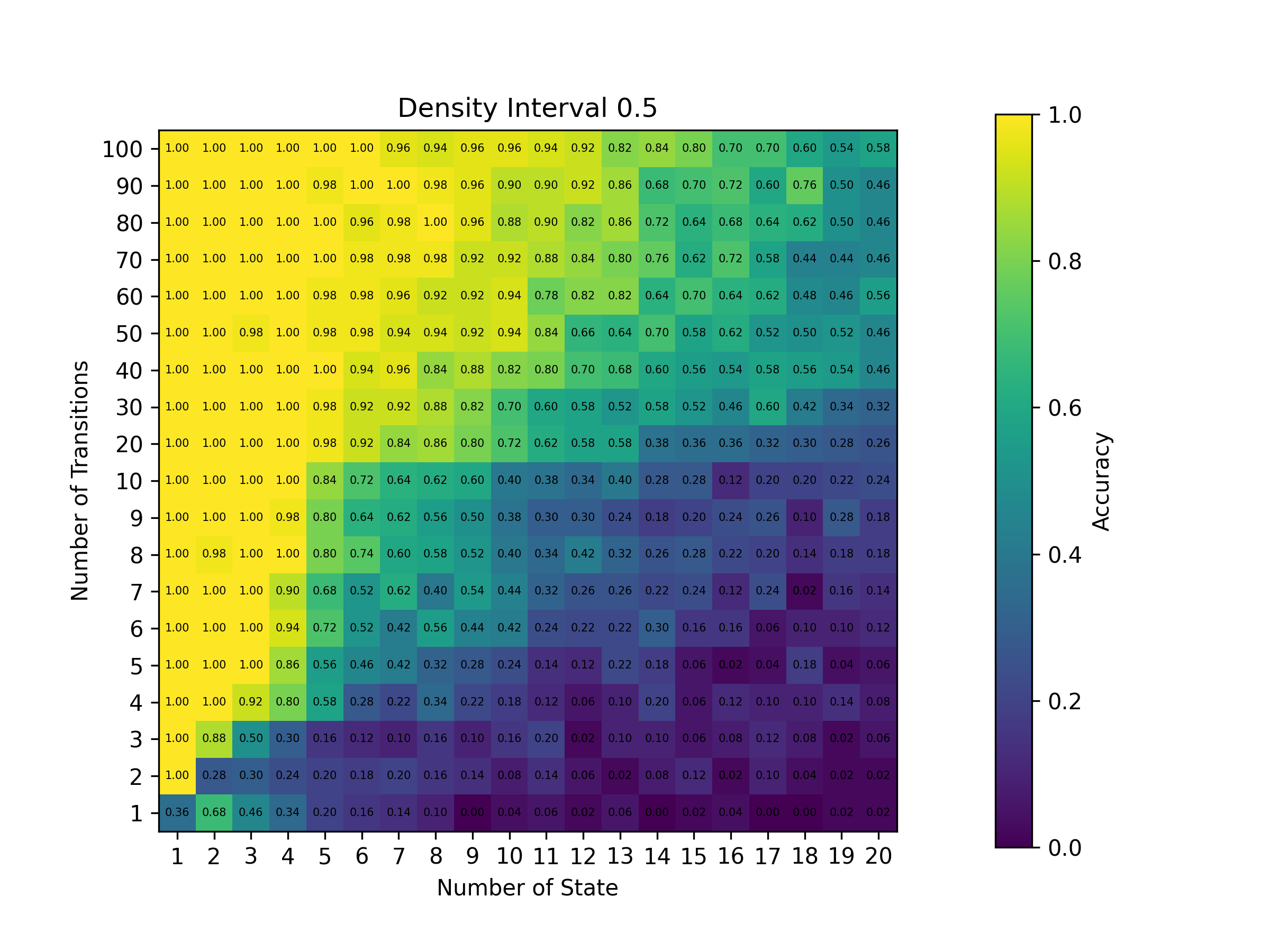}
    \includegraphics[width=0.24\linewidth]{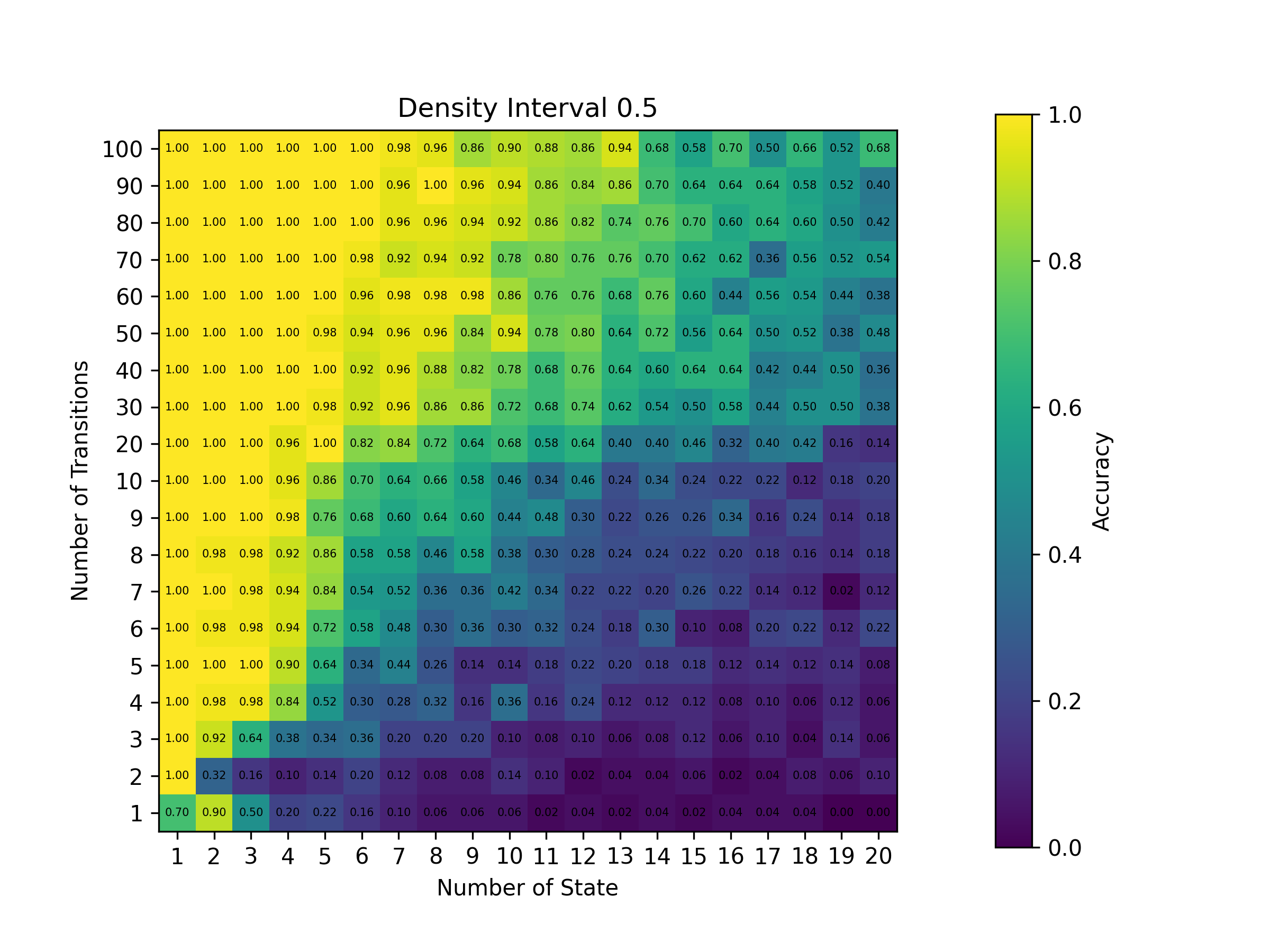}
    \caption{GPT2 Model Performance (Small, Medium, Large, X-Large)}
    \label{fig:dfa-GPT2}
\end{figure}
\begin{figure}[H]
    \centering
    \includegraphics[width=0.25\linewidth]{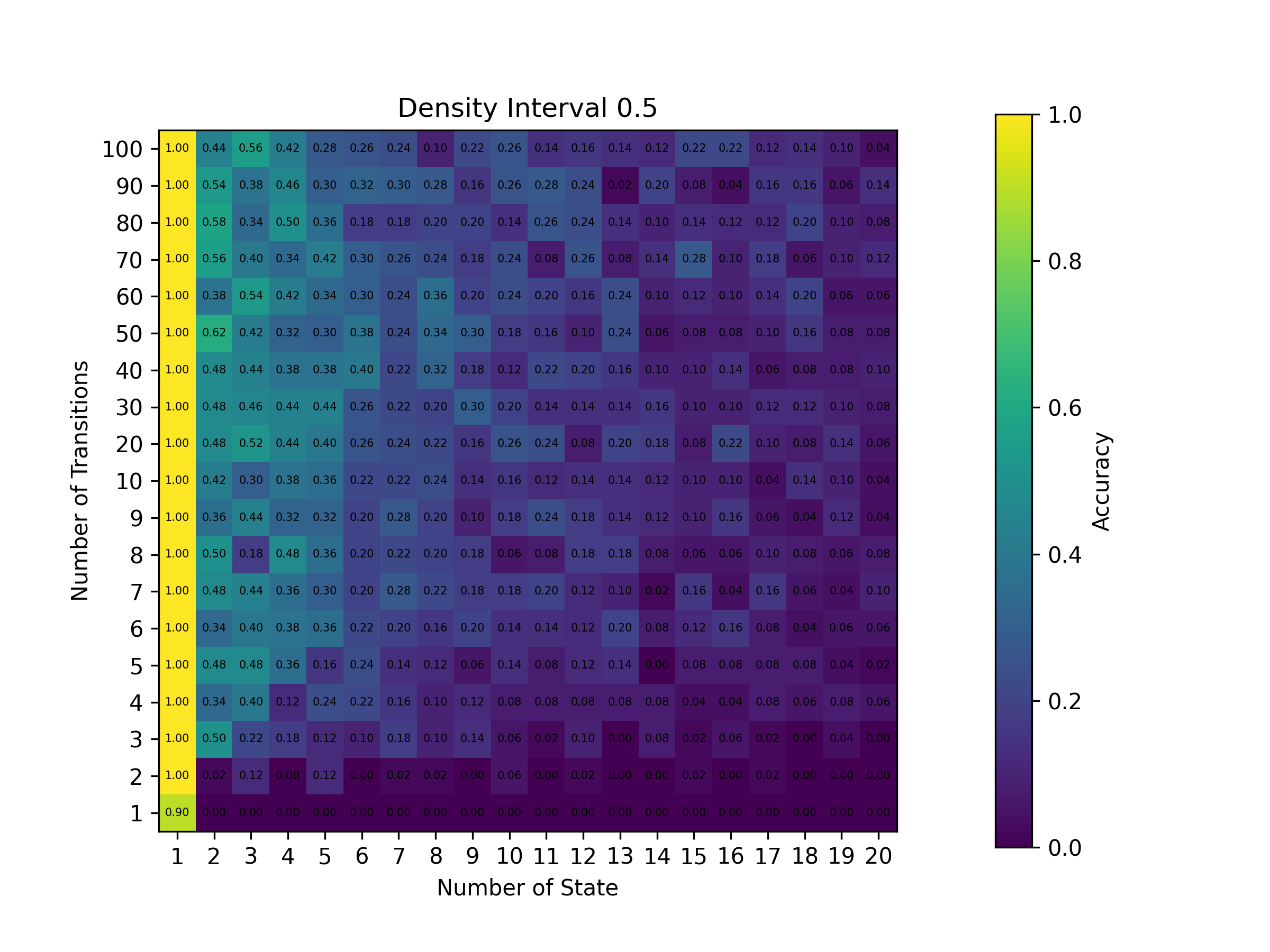}
    \includegraphics[width=0.25\linewidth]{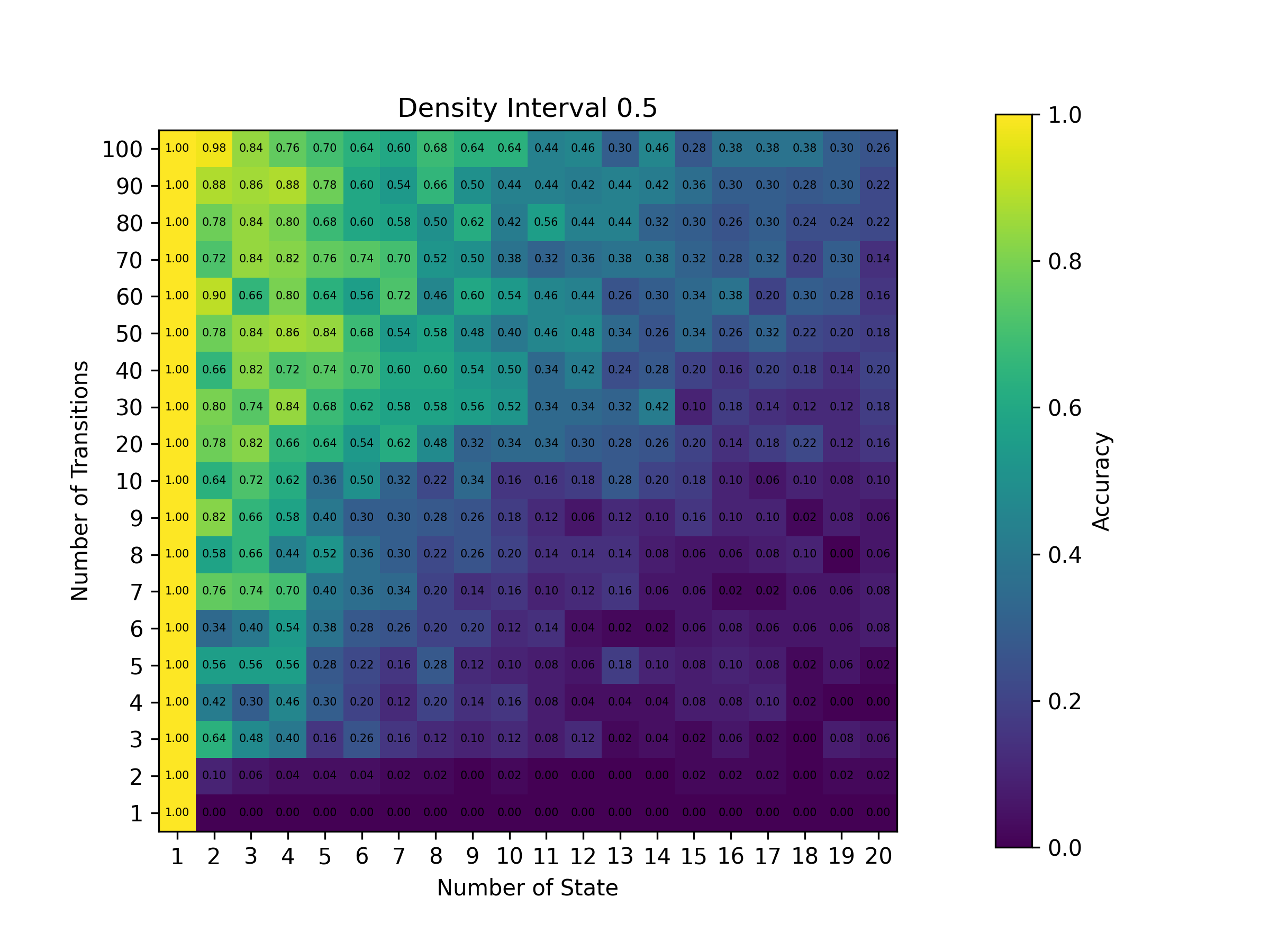}
    \includegraphics[width=0.25\linewidth]{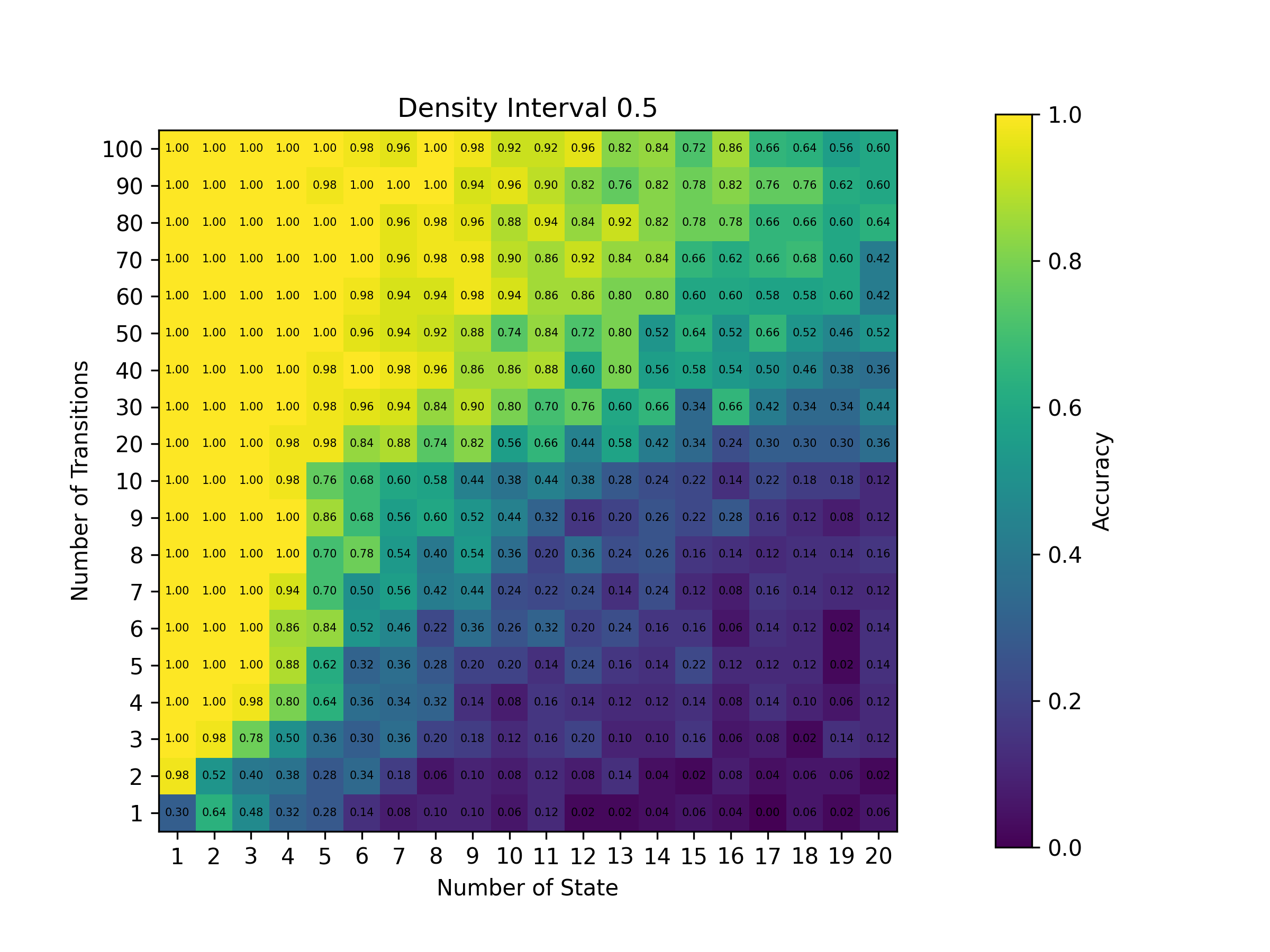}
    \caption{Pythia Model Performance (14M, 70M, 1B)}
    \label{fig:dfa-pythia}
\end{figure}

The results show emergence of state-dynamics tracking depends on LLM scale,  lower-bounded by parameter count or mandates certain training-data distribution e.g. though TinyStories 33M is $>2\times$ param count of Pythia 14M the model's performance is demonstrably worst of all models -- showing isolated pockets of nonzero accuracy. Generally, all models achieve perfect accuracy in low-state large-transition regimes; accuracy declines as the number of states increases and generally improves with more transitions. We understand whether improvements reflect state-tracking or simple pattern matching, with patching and mechanistic analysis. Models of all scales struggle with large-state low-transitions, where the LLM has not seen enough of the state space to infer transition dynamics. 

GPT2 models generally show more stable scaling and higher resilience to DFA complexity, with smoother degradation in accuracy for larger states as model size increases. Pythia models appear more sensitive to DFA state-size, showing sharper degradation in accuracy ;this degradation is faster for smaller model sizes. 

Using the experimental approach in \ref{subsec:explained_experiments}, we compute the average patching metric over $100$ clean/corrupted pairs ranging across $[1,10] \cup [20,100]$ transitions. The results are shown across \ref{app:seq-tinystories-33m} to \ref{app:head-pythia-1b}; the different-state same-action counterfactual reveals that relevant modified state information is not reliably propagated to the final token, and more notably, action information is also not moved—indicating that joint state-action reasoning is not preserved. However, in contrast, patching over irrelevant actions shows clear movement of state information over long irrelevant sequence to the final token, often occurring at Layer 24 for GPT2-XL and at Layer 11 for Pythia-1B. Across both patching experiments, information propagation occurs at these layers and the LLMs fail to route action information to the last token. For head-level patching, we find both (state-action pair and state-history) behaviors we patch for rely on the same subset of heads/layers for both positive state propagation and suppression of irrelevant signals. For instance, GPT2-XL generally uses Head 20 of Layer 22 while Pythia-1B strongly relies on Layer 14, Head 2; Layer 12, Head 6; Layer 10, Head 6; and Layer 11, Head 2.

Further analysis of attention heads in \ref{app:a12},\ref{app:a13} and \ref{app:a14} shows that GPT-XL Layer 22 Head 20 and Pythia-1B Layer 6 Head 12 are both next-token heads. Overall, evidence of state tracking is most-likely limited to pattern matching (without attention across state-action history) in the DFA domain; the behavior seems to emerge from weak interaction of several next token heads.

\subsection{Complex Text Game Task}

To evaluate the state performance of the model in realistic scenarios using a more complex language, we designed a simple real-world scenario involving $n$ number of entities entering a fruit store. Provided $[1, x < n ]$ number of clues for each prompt that assign $n$ fruits to each entity using natural language only, the model must predict which fruits a remaining entity is able to receive. As the clues increase, the possible receivable fruit choices decrease. The accuracy plots related to this task is pictured below for each model:

\begin{figure}[H]
    \centering
    \includegraphics[width=0.24\linewidth]{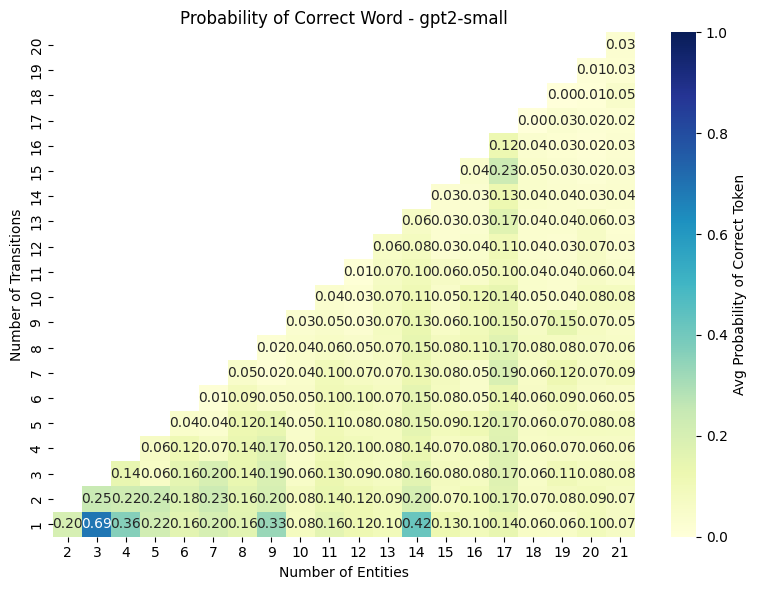}
    \includegraphics[width=0.24\linewidth]{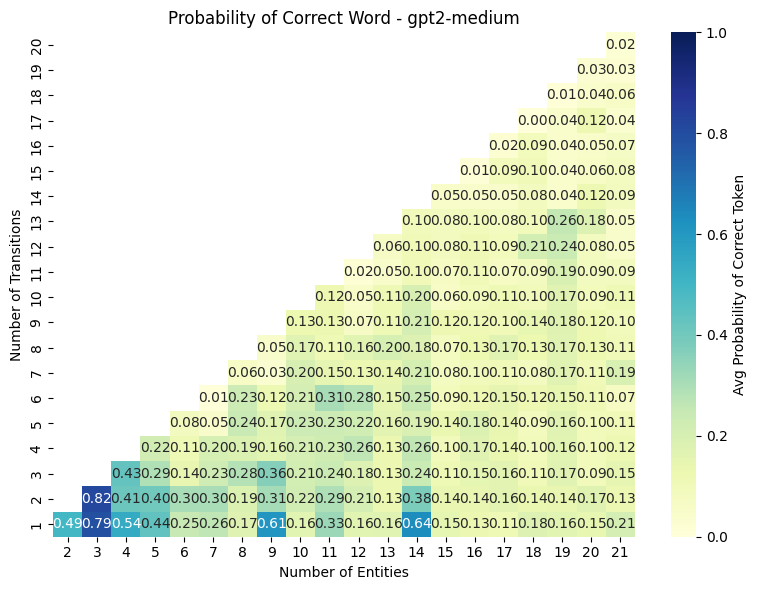}
    \includegraphics[width=0.24\linewidth]{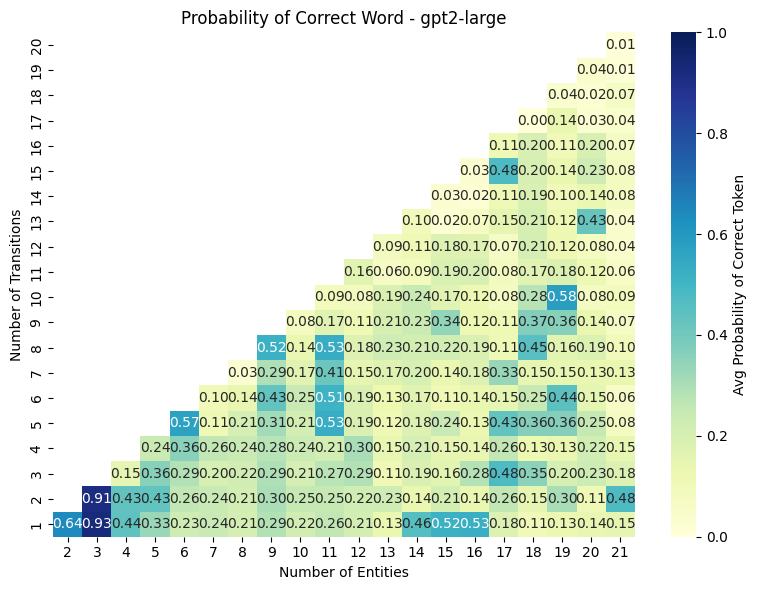}
    \includegraphics[width=0.24\linewidth]{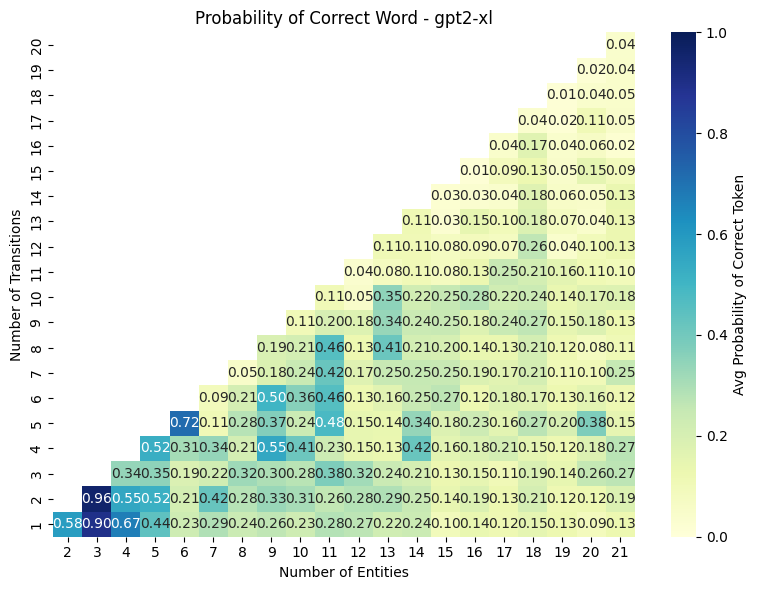}
    \caption{GPT2 Model Performance (Small, Medium, Large, X-Large)}
    \label{fig:dfa-GPT2}
\end{figure}
\begin{figure}[H]
    \centering
    \includegraphics[width=0.25\linewidth]{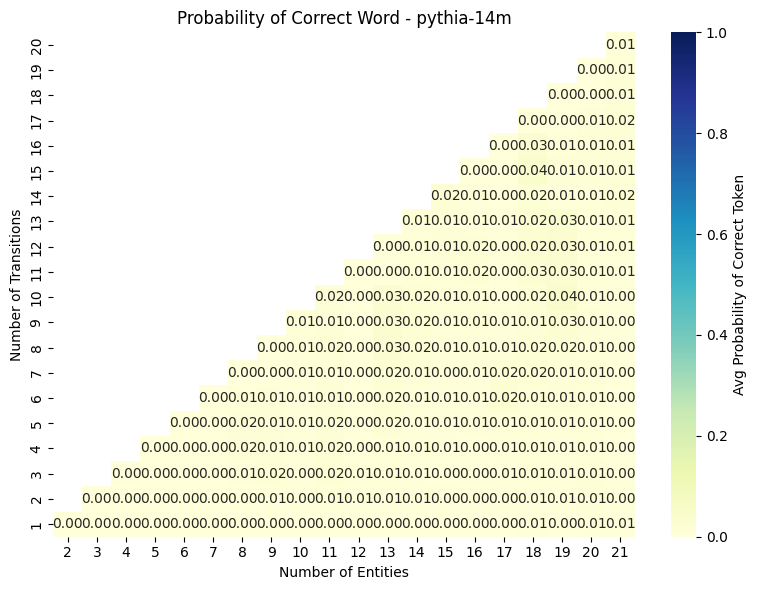}
    \includegraphics[width=0.25\linewidth]{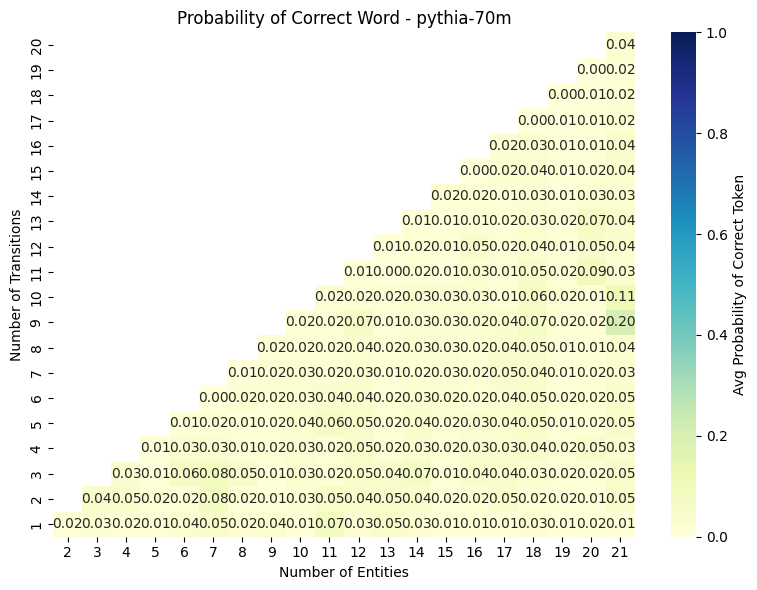}
    \includegraphics[width=0.25\linewidth]{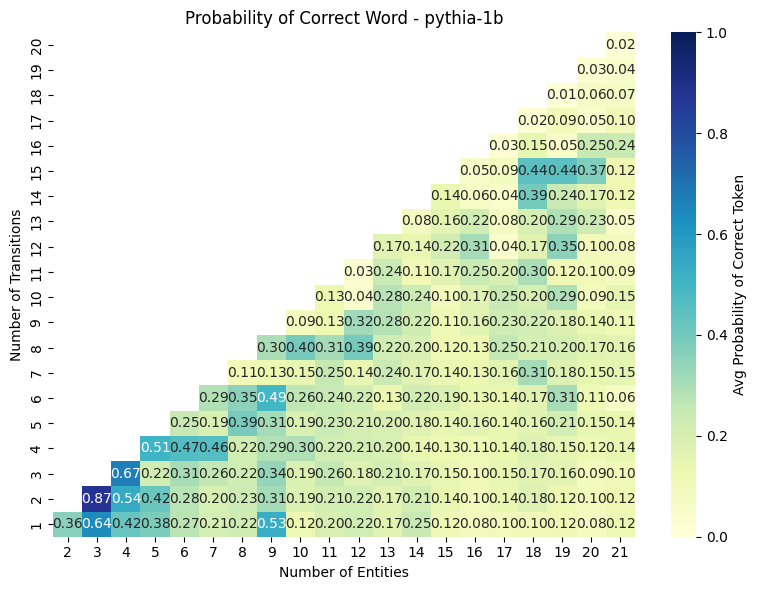}
    \caption{Pythia Model Performance (14M, 70M, 1B)}
    \label{fig:dfa-pythia}
\end{figure}
\begin{figure}[H]
    \centering
    \includegraphics[width=0.25\linewidth]{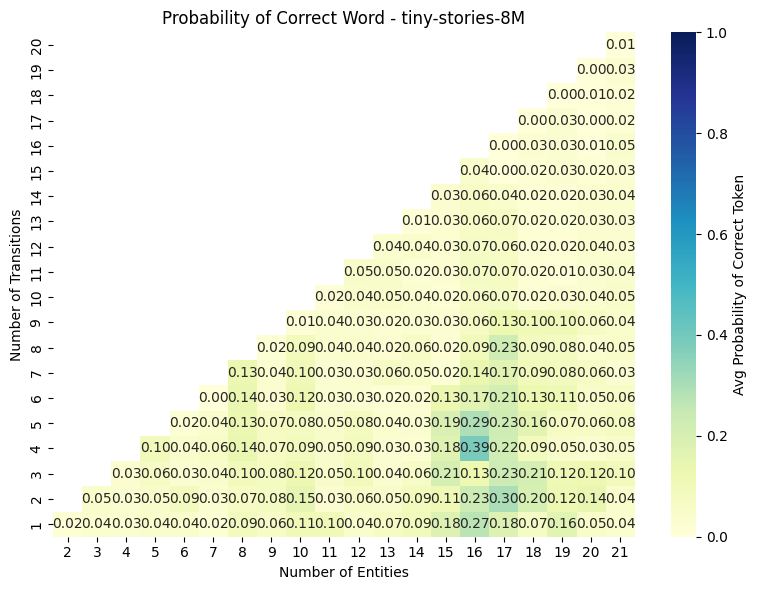}
    \includegraphics[width=0.25\linewidth]{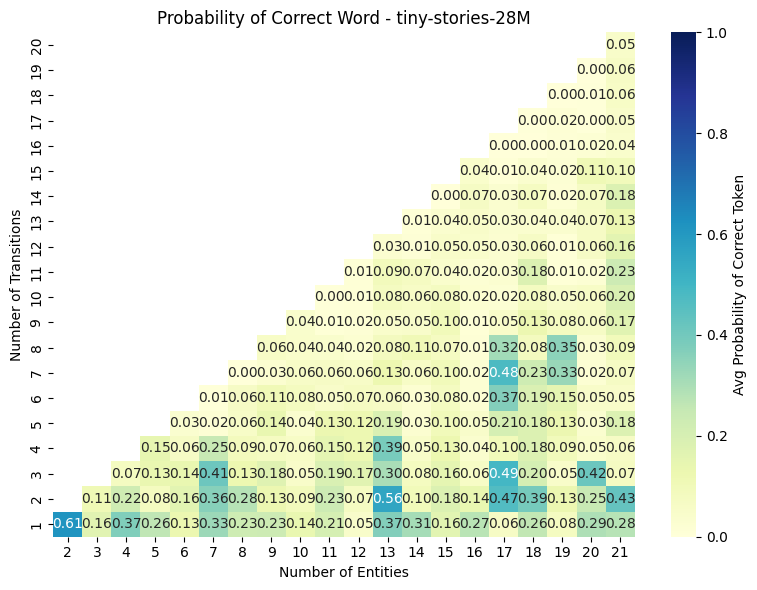}
    \includegraphics[width=0.25\linewidth]{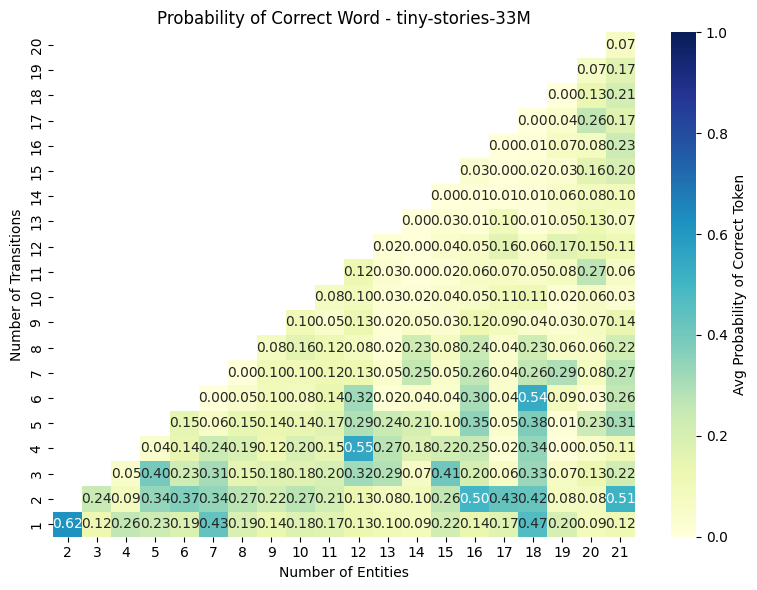}
    \caption{Tiny Stories Model Performance (8M, 24M, 33M)}
    \label{fig:dfa-pythia}
\end{figure}

Performance on this task was substantially lower than in the previous two tasks, with models generally achieving higher accuracy when $n \leq 15$ and the number of clues was moderate. The largest models performed best overall, with GPT-2 XL and Pythia 1B showing comparable results. These findings align with our hypothesis: the increased linguistic complexity and combinatorial reasoning required by the natural language clues impose a significant challenge on the model’s predictive ability. Contributing factors likely include linguistic noise, punctuation variability, inconsistent scenario framing, and limited token co-occurrence between proper names and fruit labels.

To further investigate internal model behavior, we performed residual stream activation patching in 20 clean-corrupt prompt pairs. We observed heightened attention to tokens corresponding to, or adjacent to, fruit words. Since the patched tokens varied across prompt pairs, this outcome was expected when averaged across trials. Consistently, the highest logit difference was localized at the final token of the prompt across all test cases, indicating that the model integrates contextual information throughout the sequence to inform its final prediction. Additionally, we found disproportionately high attention contributions from Head 21 at Layer 6, Head 24 at Layer 14, and Head 26 at Layer 14.

\begin{figure}[!ht]
    \centering
    \begin{subfigure}[t]{0.8\textwidth}
        \centering
        \includegraphics[height=4.5cm]{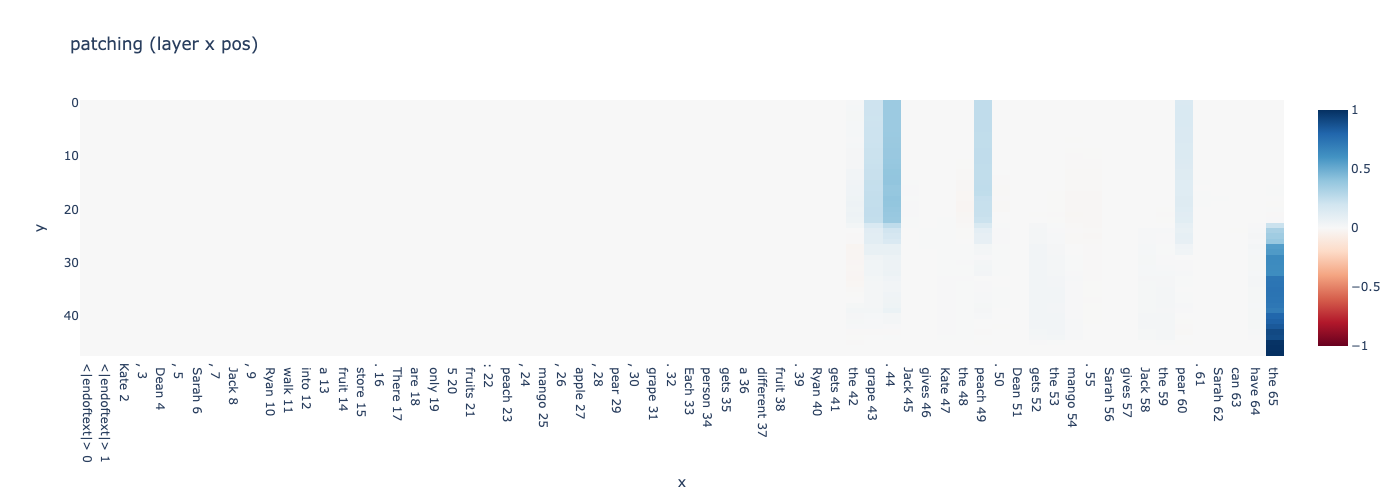}
        \caption{Activation patching of residual stream with GPT-2 XL}
        \label{fig:patching_layer_pos_heatmap_gptxl}
    \end{subfigure}%
    \hfill
    \begin{subfigure}[t]{0.49\textwidth}
        \centering
        \includegraphics[height=4.5cm]{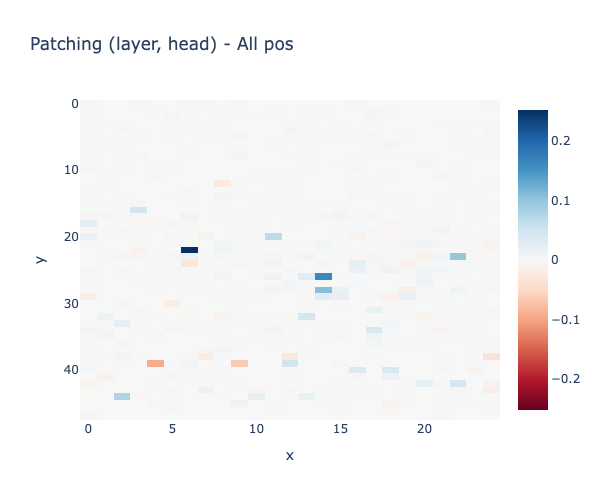}
        \caption{Activation patching for GPT-2 XL on the Fruit Store task}
        \label{fig:patching_heatmap_gptxl}
    \end{subfigure}
    \caption{Comparison of activation patching results for GPT-2 XL. Left: residual stream patching across token positions. Right: attention head patching.}
    \label{fig:combined_patching_heatmaps}
\end{figure}
\section{Conclusion}

Common behavior has been observed throughout the three setups, giving us an insight into the inner-workings of these Large Language Models when given state-tracking problems. This can be summarized in two phases - Induction Head formation and Induction Head optimization. 

In the earlier layers of the models, multiple Induction Heads form, with each attending to one, all, or subsets of previous occurrences of states (such as the A, B, and C states in the Box experiment). Then, Induction Head optimization occurs. This happens through the formation of mid-layer Induction Heads that either enhance these previous heads, or inhibit them.

We have also noticed Name-Mover Heads that attempt to associate the state with the right object, such as "watch" to its corresponding "A" state. These heads do not always form effectively, yet larger models appear to increase its formation effectiveness. 

In summary, our results indicate that state-tracking is not a behavior led by an individual attention head. Rather, state-tracking appears to be the result of different types of attention heads working together.

For future directions, we encourage a deeper exploration of these Name-Mover Heads, focusing on identifying the factors aiding in their emergence. More thorough patching techniques would also be fruitful in reinforcing these results, such as path patching.

\newpage
\bibliography{proposal}
\appendix
\section{Appendix}

\subsection{Explained Experimental Procedures}
\label{subsec:explained_experiments}

\subsubsection{Attribution of state-tracking}
Our activation patching experiment involves pairs of clean and corrupted prompts. A \textit{clean prompt} is a standard scenario generated using the DFA specified for the domain, for which the model is expected to predict the correct final state/outcome. A \textit{corrupted prompt} is created by making a minimal change to a clean prompt, these vary for the different domain we evaluate:
\begin{itemize}
    \item Box Tracking: altering either the initial placement of the queried object or its last transition, such that the correct final box for the queried object changes e.g. if the queried object ``watch'' was initially in ``Box A'' and later moved to ``Box B'' (clean answer ``B''), a corrupted prompt might change its last move to ``Box C'' (corrupted answer ``C'')
    \item Traditional DFA (different states, same action): given a DFA constructed such that the same action `A' can be taken from 2 different states `a',`c' and lead to 2 different outcomes `b',`d', we alter one of the prior states (e.g. $c\rightarrow a$) to break the DFA e.g. if the clean prompt followed a sequence ``a A b B c A d B a A'', a corrupted prompt might change to ``a A b B a A d B a A''. This evaluate's the LLMs ability to attend to state-action pairs
    \item Traditional DFA (long sequence of irrelevant actions): given a DFA constructed such that a certain action at a state (`A' at `a') does not change the state, we alter the starting state and perform different lengths of irrelevant actions e.g. if the clean prompt followed a sequence ``a A b B a C a C a C a A'', a corrupted prompt might change to ``b A a B b C a C a C a A''. This evaluates the LLM's capability for state-history tracking 
\end{itemize}

This metric quantifies how much the patched activation restores the logit of the original (clean) correct answer. A metric of 1 indicates full restoration of the clean answer's logit, while a metric of 0 indicates no improvement over the corrupted baseline.

When evaluating this metric, the LLM is run on both the clean and corrupted prompts. All intermediate activations for the clean prompt are cached. For the corrupted prompt, we run a forward pass where we can patch (i.e., replace) an activation at a specific model component (e.g., an attention head output at a specific layer) with the corresponding activation from the clean cache.

\subsection{Evaluation of Entity Tracking on Tinystories suite}
\label{sec:ts-box}
\begin{figure}[H]
    \centering
    \includegraphics[width=0.25\linewidth]{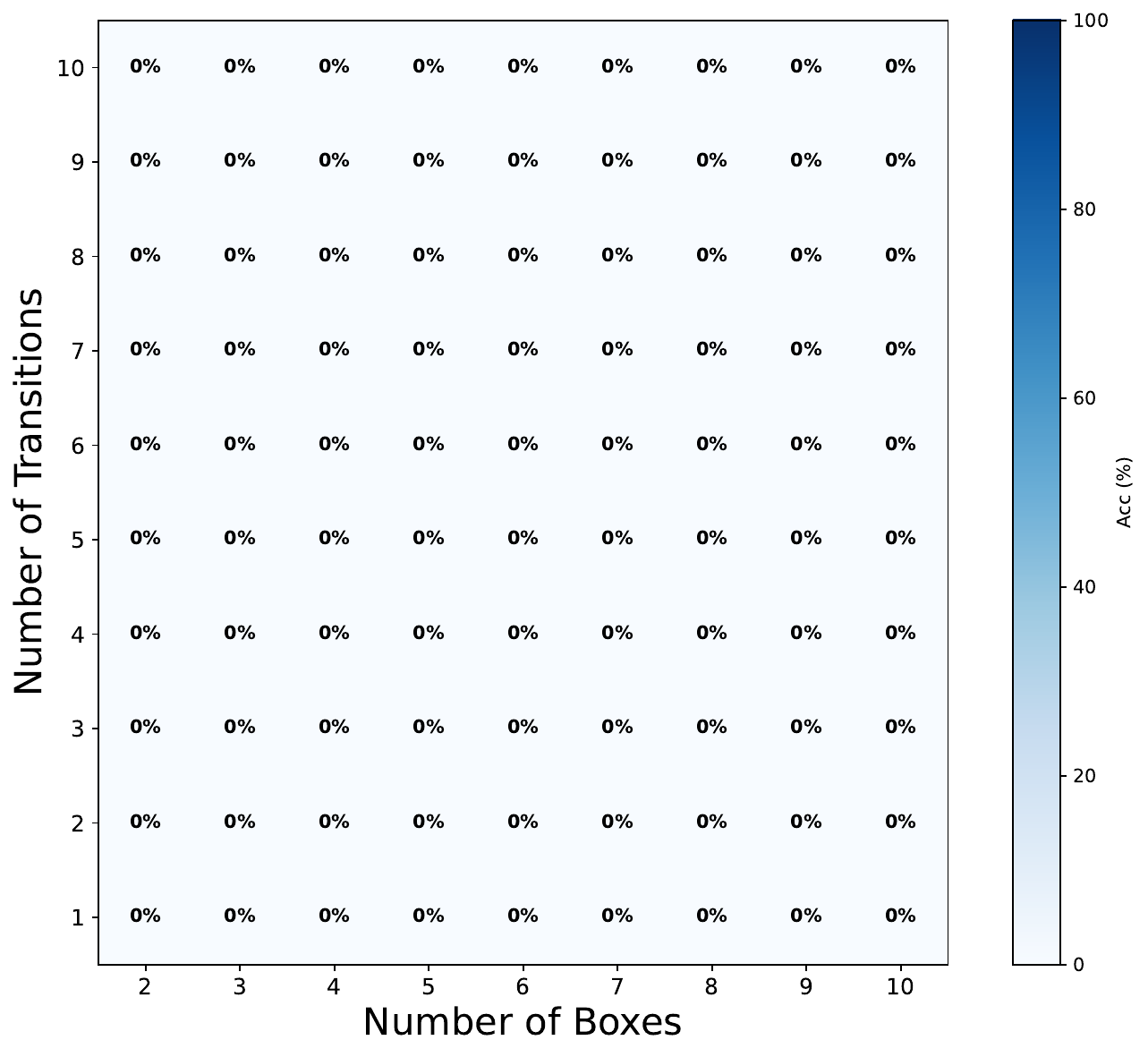}
    \includegraphics[width=0.25\linewidth]{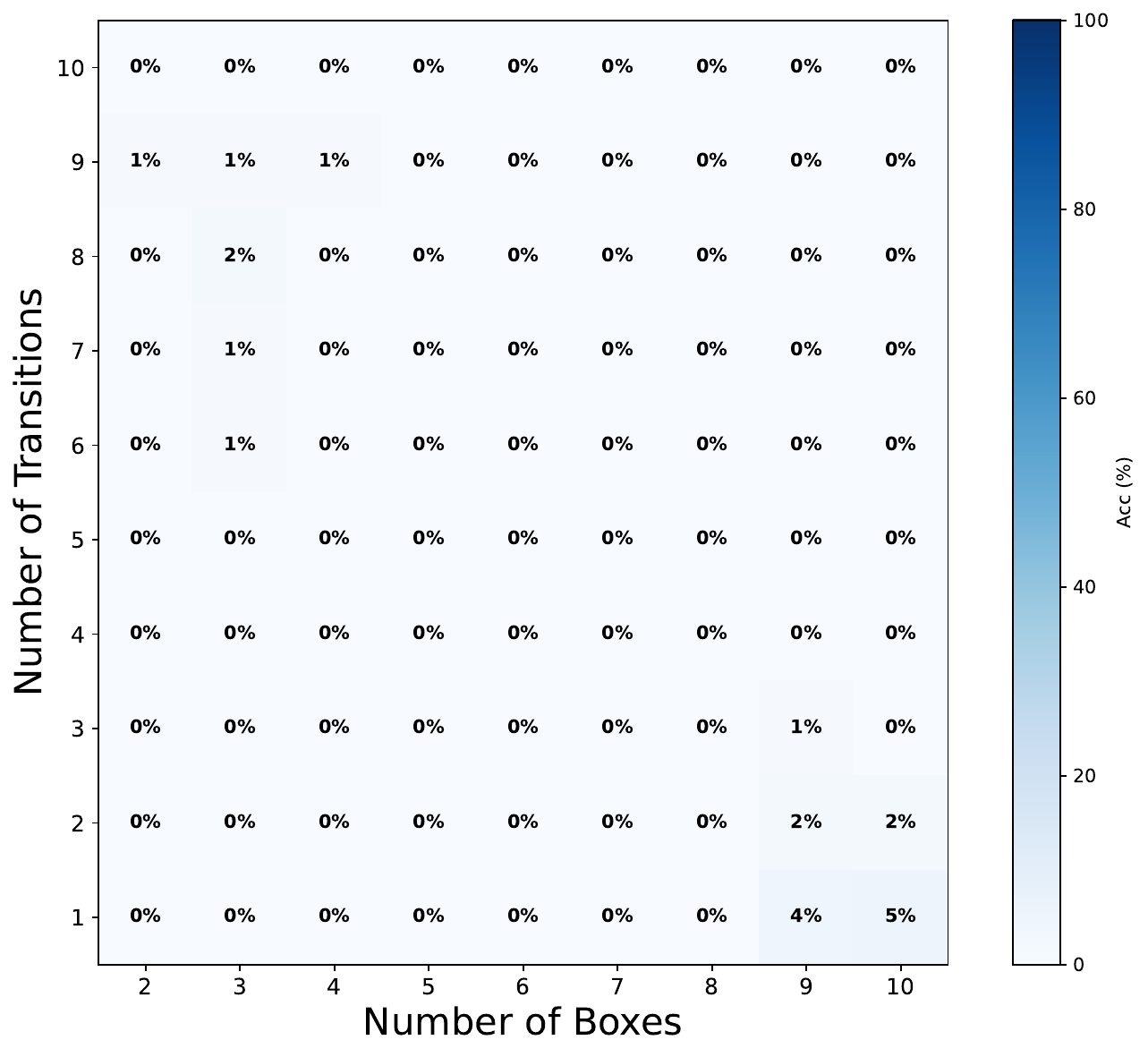}
    \includegraphics[width=0.25\linewidth]{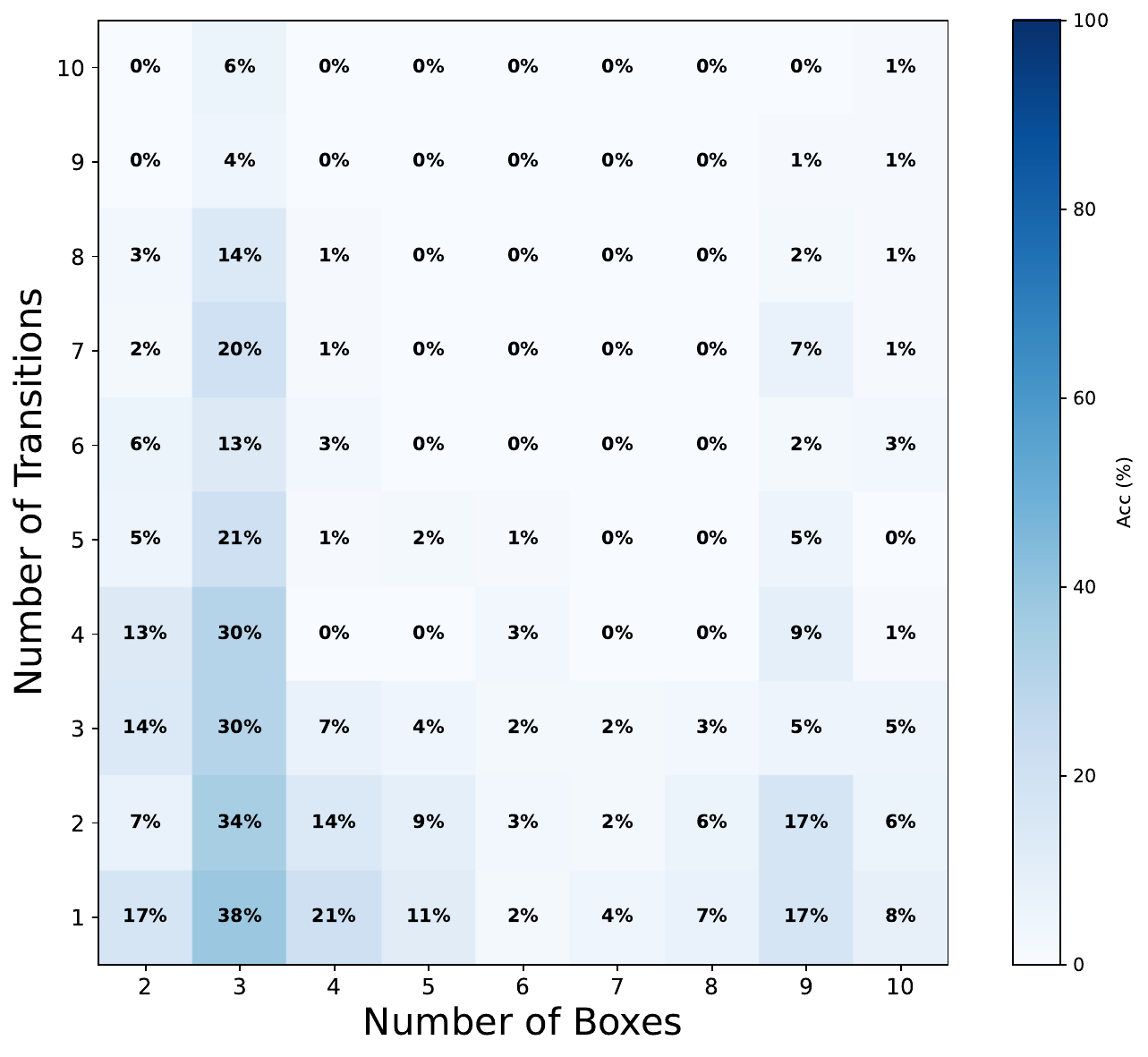}
    \caption{TinyStories Model Performance (8M, 28M, 33M)}
    \label{fig:ts-box}
\end{figure}

\subsection{Evaluation of Entity Tracking on Pythia series}
\label{sec:py-box}
\begin{figure}[H]
    \centering
    \includegraphics[width=0.25\linewidth]{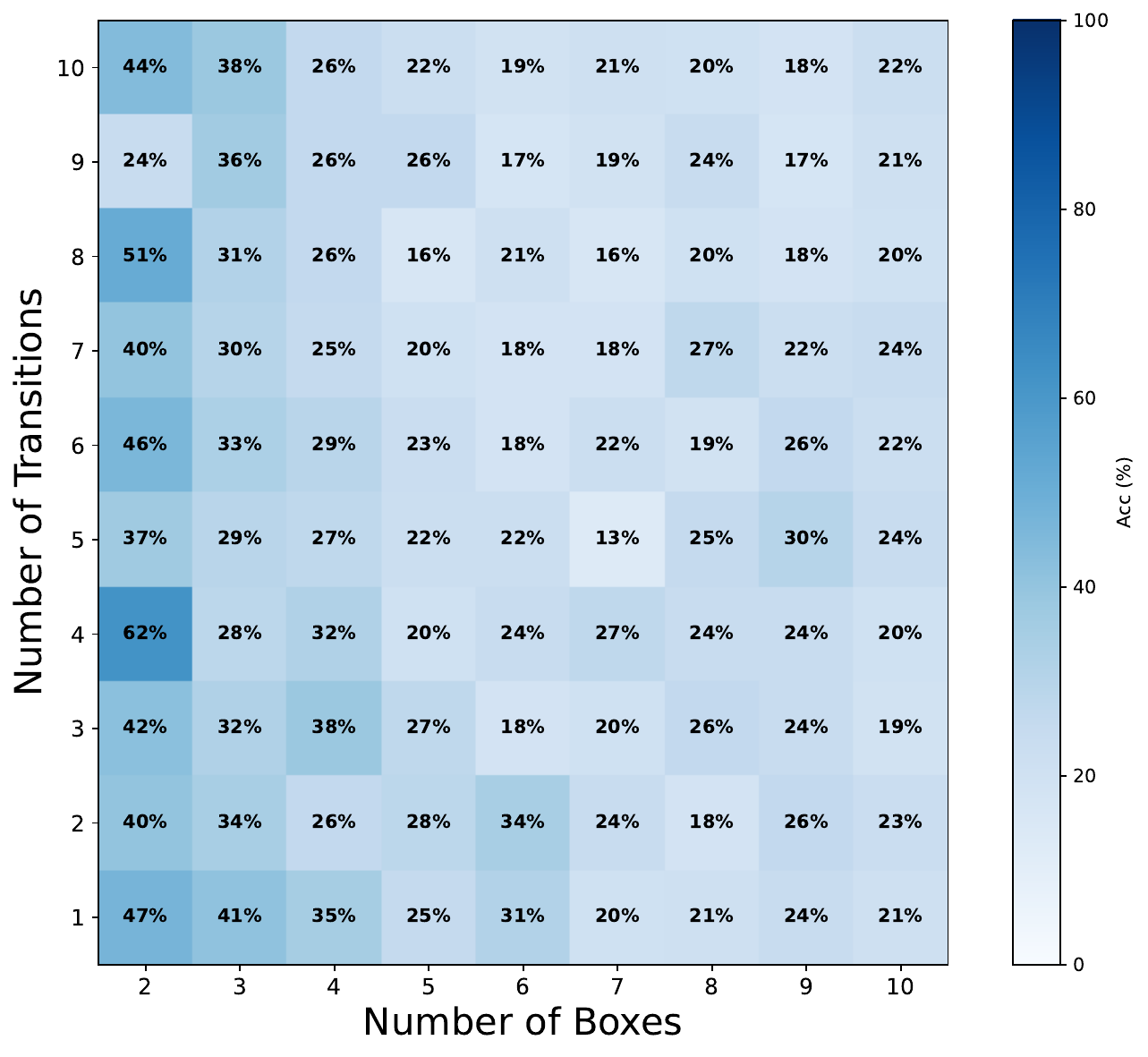}
    \includegraphics[width=0.25\linewidth]{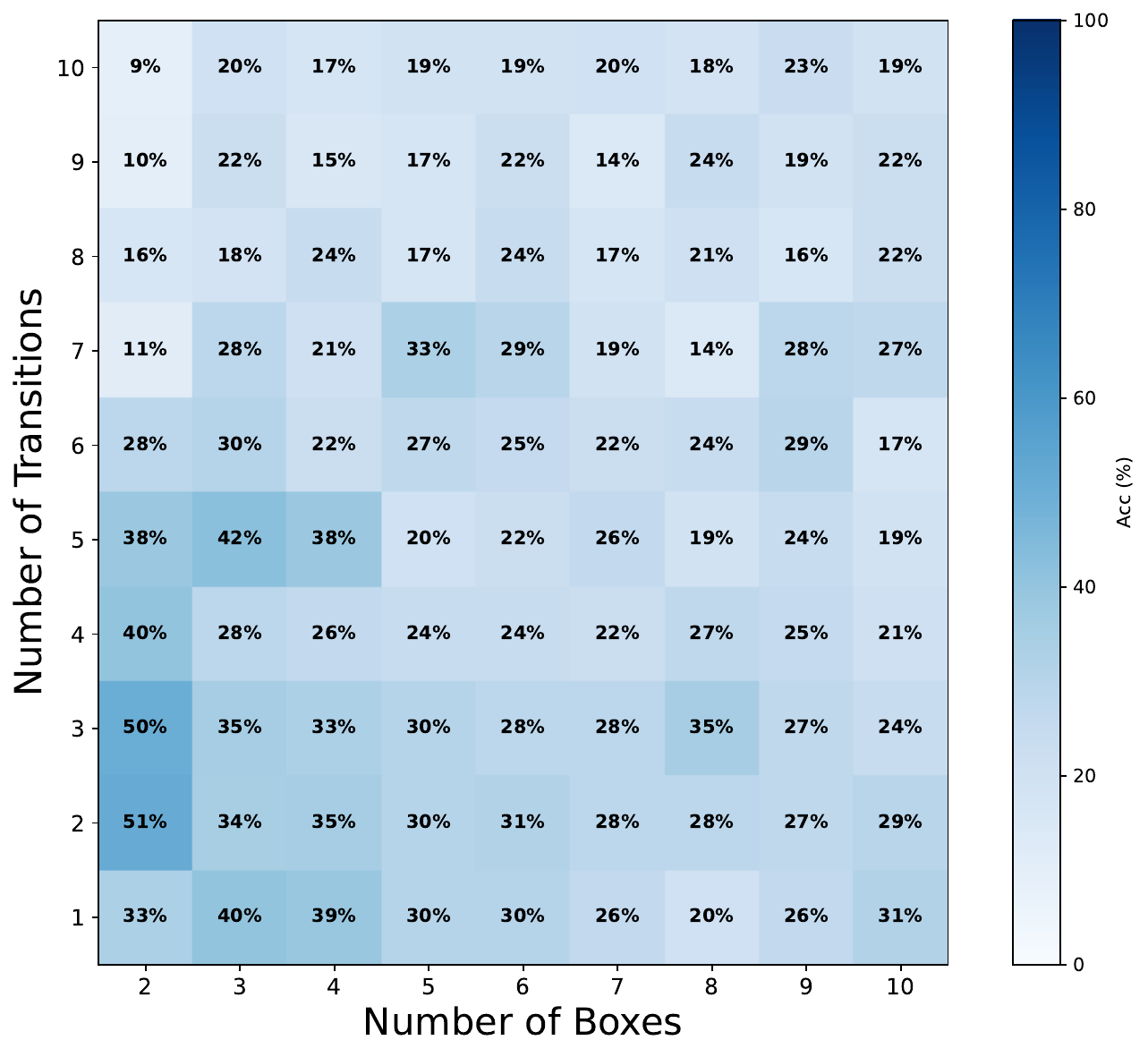}
    \includegraphics[width=0.25\linewidth]{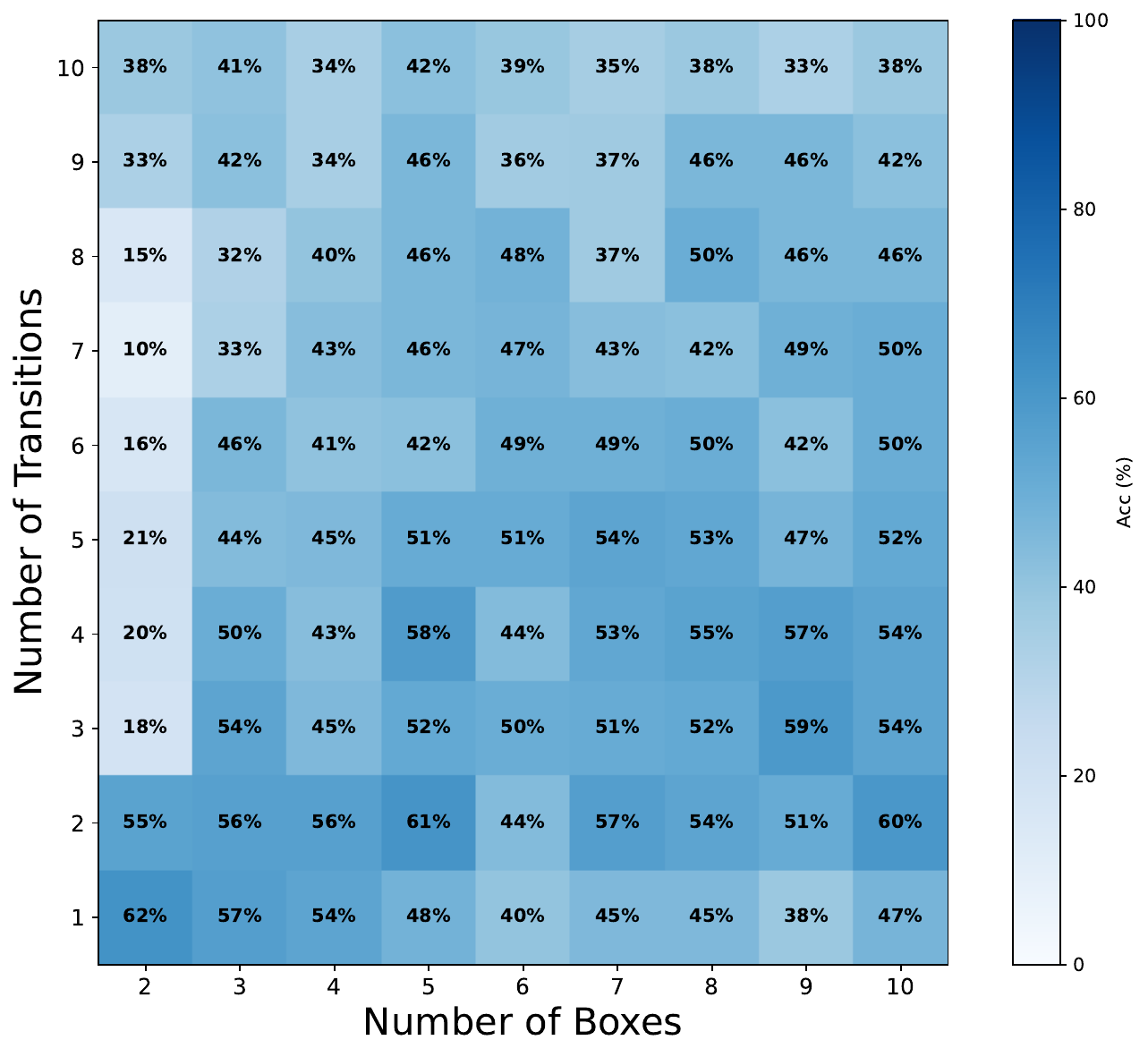}
    \caption{Pythia Model Performance (70M, 1B, 6.9B)}
    \label{fig:pythia-box}
\end{figure}

\subsection{Accuracy Line Plot: GPT2}
\begin{figure}[H]
  \centering
  \begin{subfigure}[b]{0.49\linewidth}
    \centering
    \includegraphics[width=\linewidth]{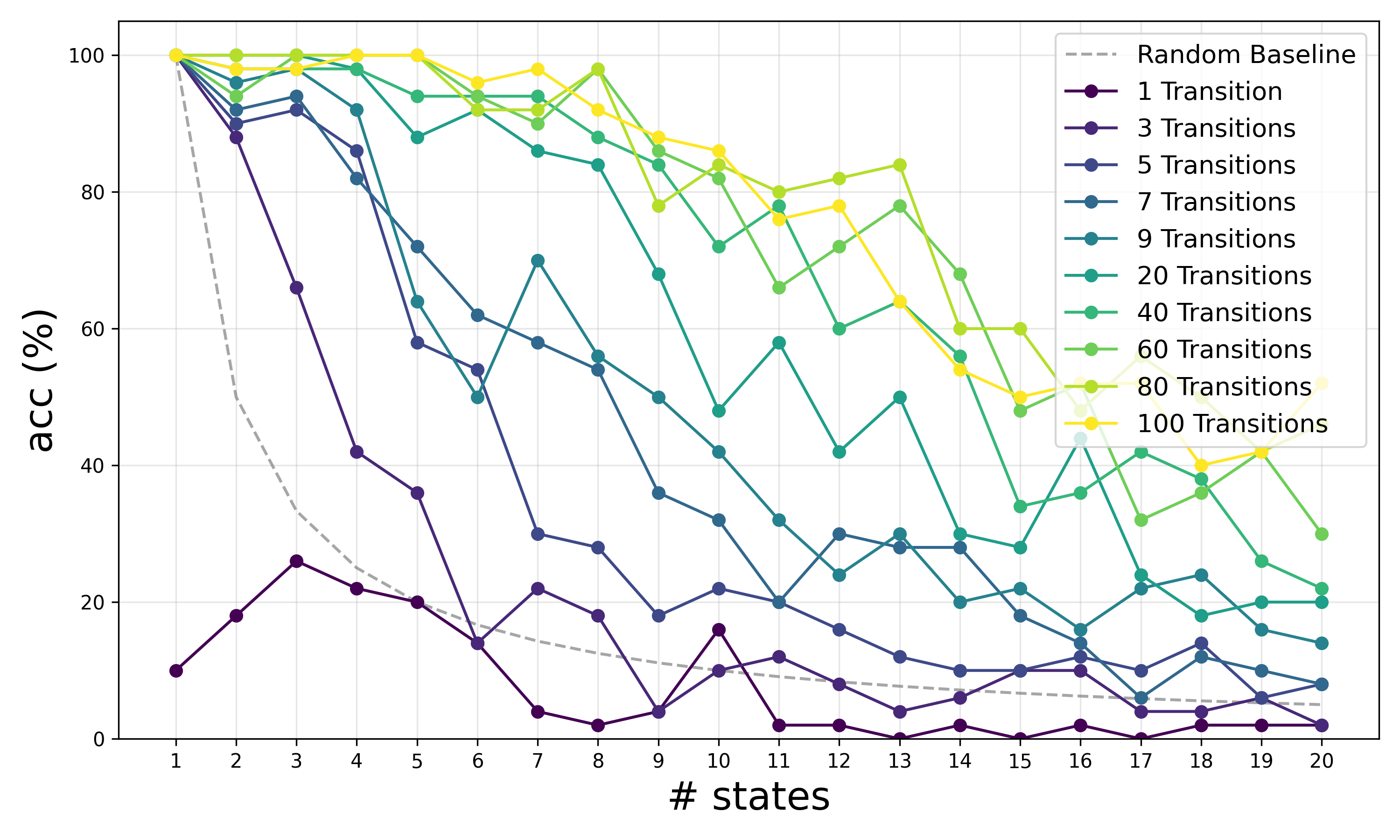}
    \caption{Small 124M}
    \label{fig:dfa-small}
  \end{subfigure}%
  \hfill
  \begin{subfigure}[b]{0.49\linewidth}
    \centering
    \includegraphics[width=\linewidth]{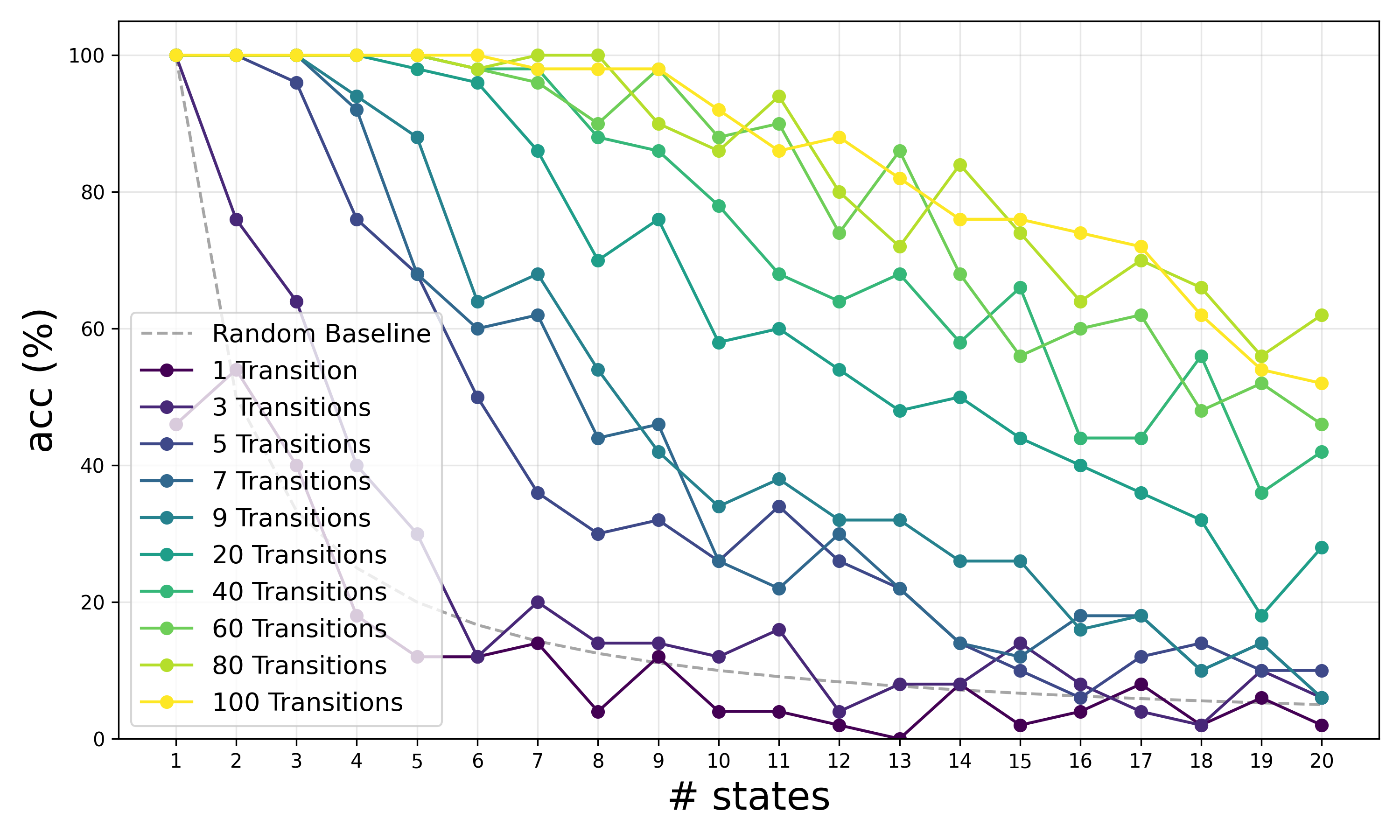}
    \caption{Medium 355M}
    \label{fig:dfa-medium}
  \end{subfigure}

  \vspace{1em}

  \begin{subfigure}[b]{0.49\linewidth}
    \centering
    \includegraphics[width=\linewidth]{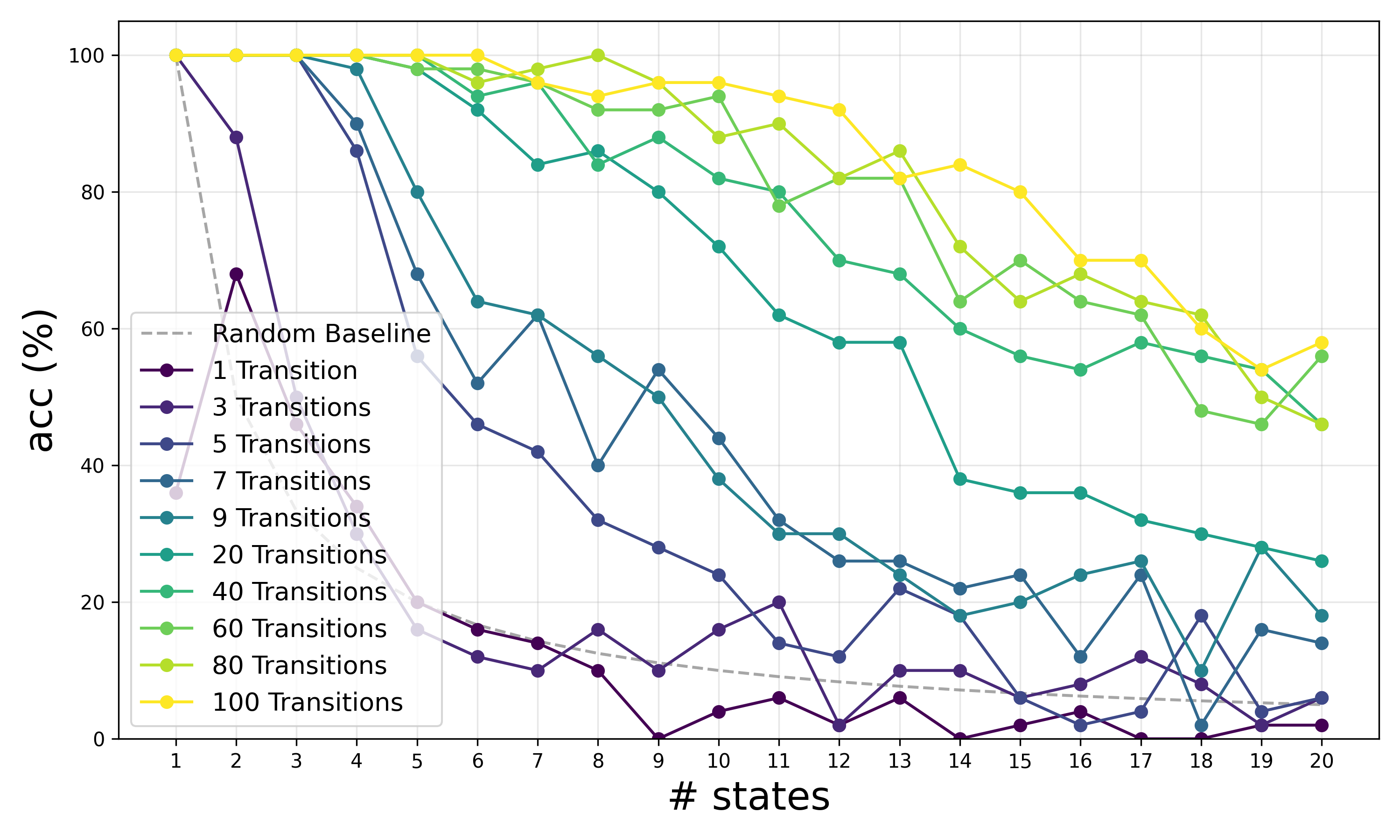}
    \caption{Large 774M}
    \label{fig:dfa-large}
  \end{subfigure}%
  \hfill
  \begin{subfigure}[b]{0.49\linewidth}
    \centering
    \includegraphics[width=\linewidth]{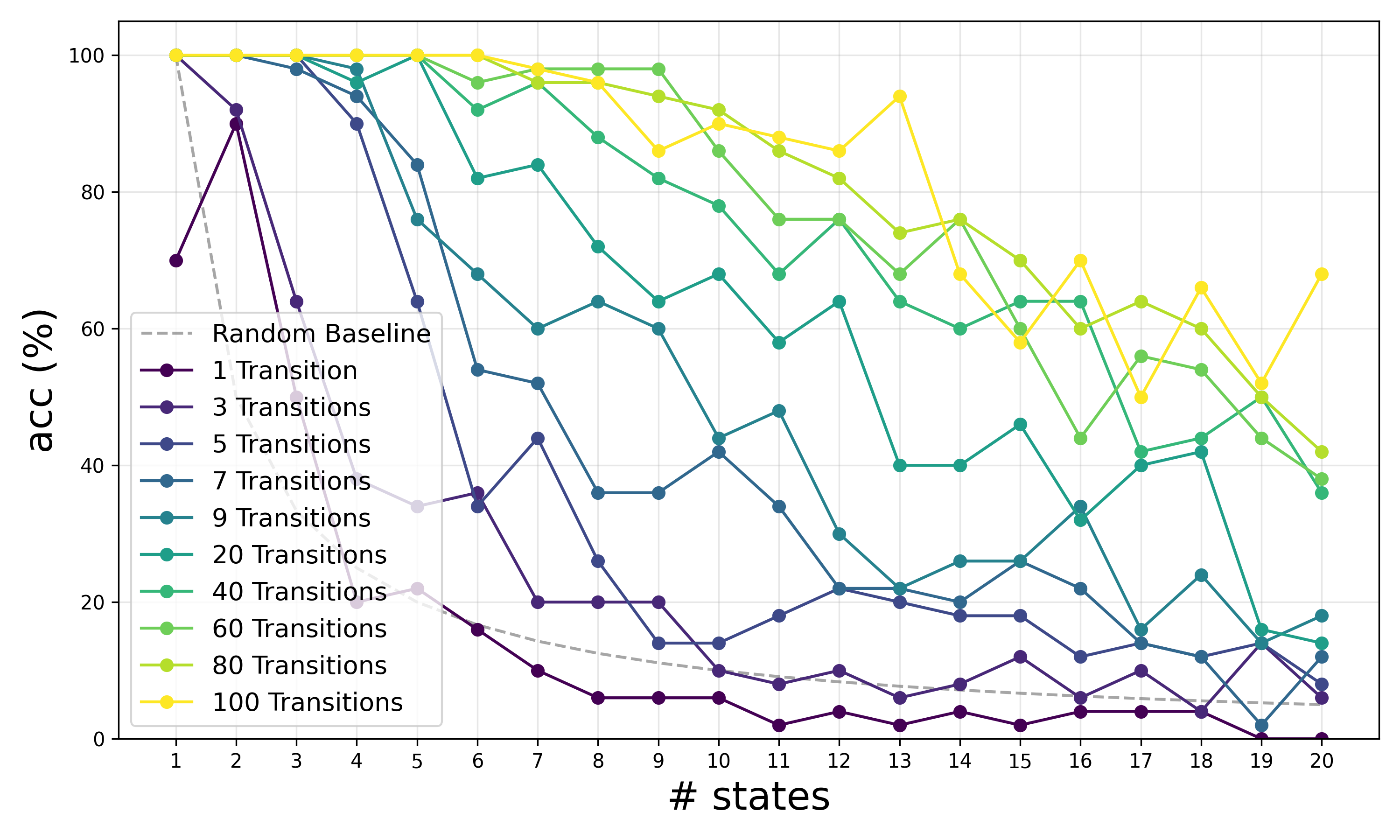}
    \caption{XL 1.5B}
    \label{fig:dfa-xl}
  \end{subfigure}

  \caption{GPT-2 DFA State–Action Accuracy by Model Size}
  \label{fig:dfa-accuracy-grid}
\end{figure}

\subsection{Sequence Patching Heatmap: TinyStories-33M}
\label{app:seq-tinystories-33m}
\begin{figure}[H]
  \centering
  \begin{subfigure}[b]{0.49\linewidth}
    \centering
    \includegraphics[width=\linewidth]{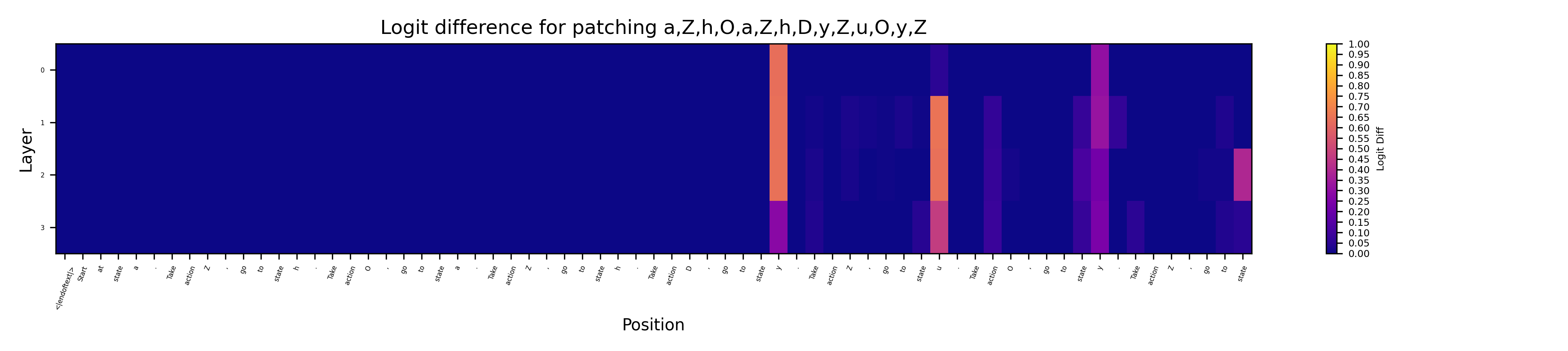}
    \caption{Different States Same Action: 6 Transitions}
    \label{fig:dfa-tinystories-2s1a-6}
  \end{subfigure}%
  \hfill
  \begin{subfigure}[b]{0.49\linewidth}
    \centering
    \includegraphics[width=\linewidth]{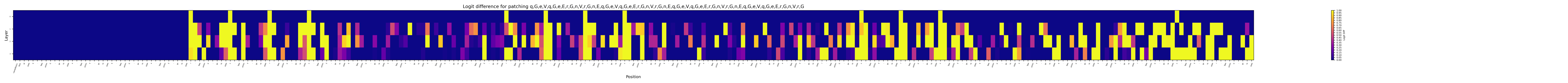}
    \caption{Different States Same Action: 30 Transitions}
    \label{fig:dfa-tinystories-2s1a-30}
  \end{subfigure}

  \vspace{1em}

  \begin{subfigure}[b]{0.49\linewidth}
    \centering
    \includegraphics[width=\linewidth]{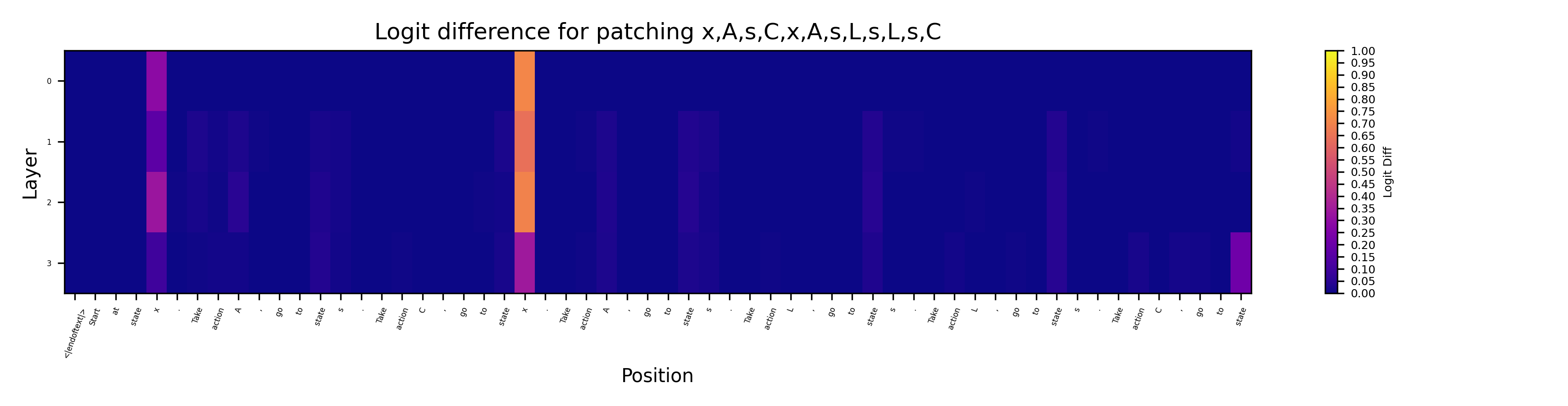}
    \caption{Long Sequences of Irrelevant Actions: 5 Transitions}
    \label{fig:dfa-tinystories-noop-5}
  \end{subfigure}%
  \hfill
  \begin{subfigure}[b]{0.49\linewidth}
    \centering
    \includegraphics[width=\linewidth]{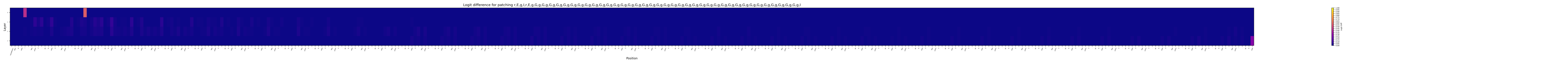}
    \caption{Long Sequences of Irrelevant Actions: 40 Transitions}
    \label{fig:dfa-tinystories-noop-40}
  \end{subfigure}

  \caption{TinyStories-33M Sequence Patching Heatmap}
  \label{fig:dfa-tinystories-sequencepatching-grid}
\end{figure}

\subsection{Head Patching Heatmap: TinyStories-33M}
\label{app:head-tinystories-33m}
\begin{figure}[H]  \centering
  \begin{subfigure}[b]{0.49\linewidth}
    \centering
    \includegraphics[width=\linewidth]{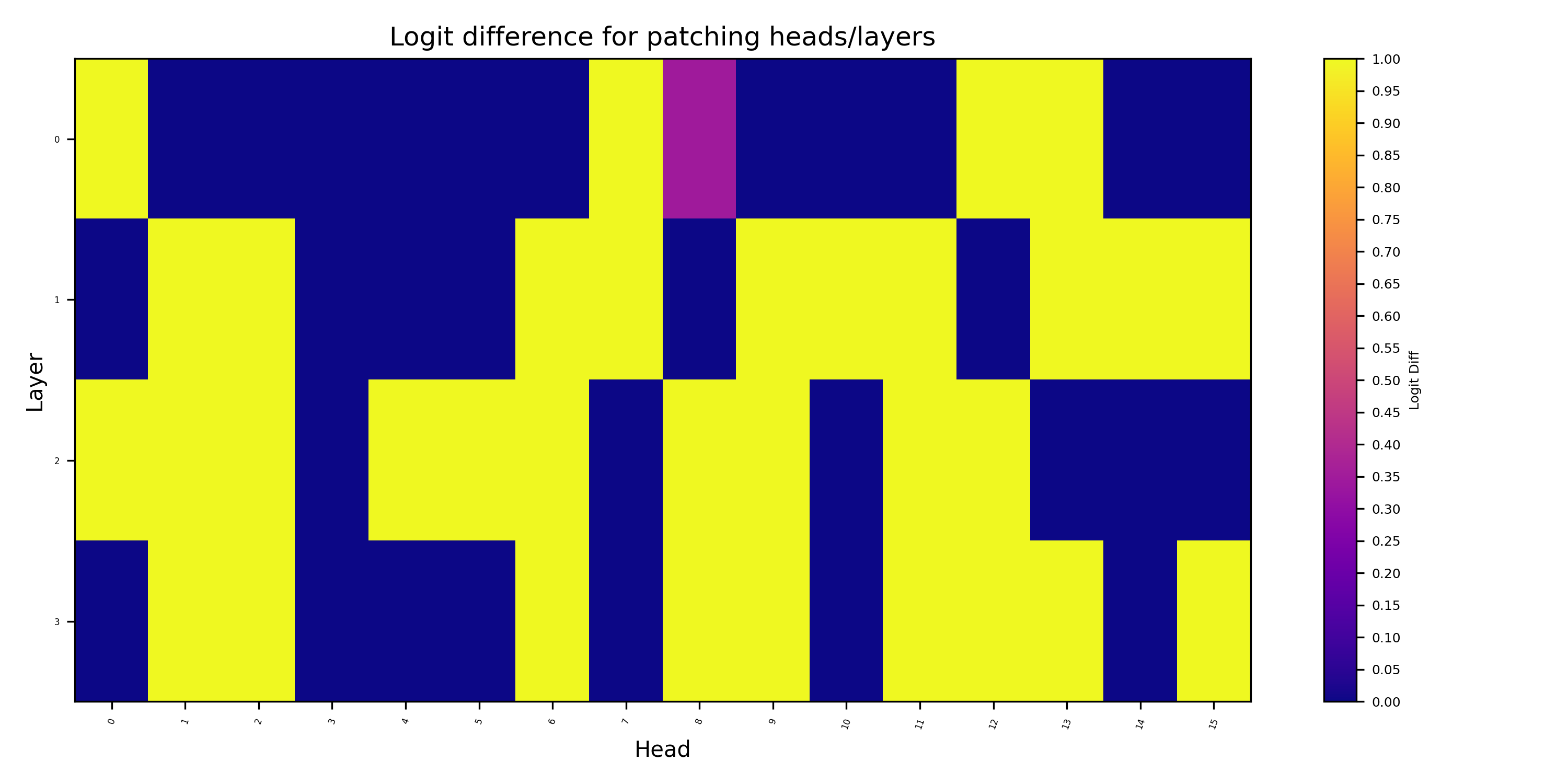}
    \caption{Different States Same Action}
    \label{fig:dfa-tinystories-2s1a-head}
  \end{subfigure}%
  \hfill
  \begin{subfigure}[b]{0.49\linewidth}
    \centering
    \includegraphics[width=\linewidth]{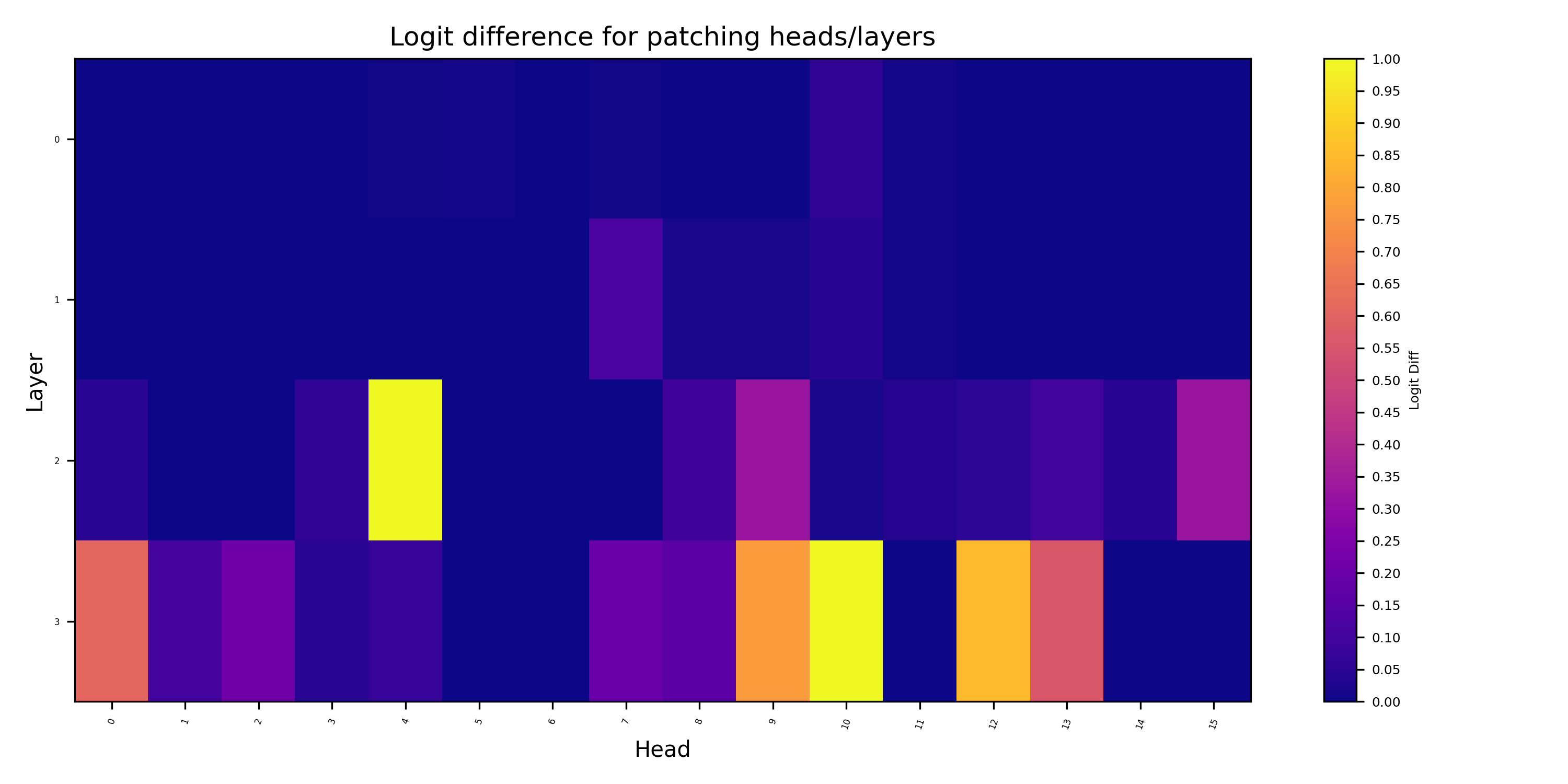}
    \caption{Long Sequences of Irrelevant Actions}
    \label{fig:dfa-tinystories-noop-head}
  \end{subfigure}

  \caption{TinyStories-33M Head Patching Heatmap}
  \label{fig:dfa-tinystories-headpatching-grid}
\end{figure}

\subsection{Sequence Patching Heatmap: GPT2-XL}
\label{app:seq-gpt2-xl}
\begin{figure}[H]
  \centering
  \begin{subfigure}[b]{0.49\linewidth}
    \centering
    \includegraphics[width=\linewidth]{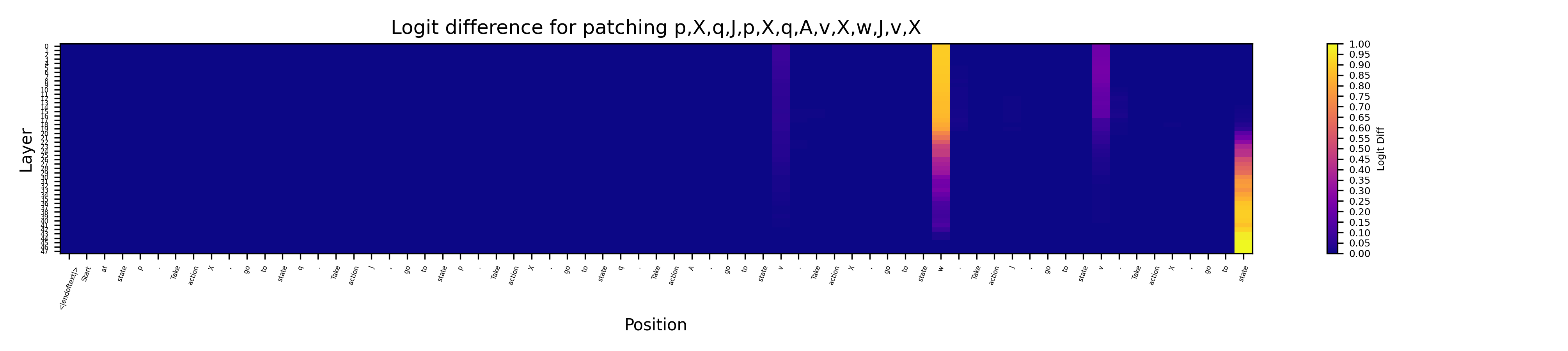}
    \caption{Different States Same Action: 6 Transitions}
    \label{fig:dfa-gpt-2s1a-6}
  \end{subfigure}%
  \hfill
  \begin{subfigure}[b]{0.49\linewidth}
    \centering
    \includegraphics[width=\linewidth]{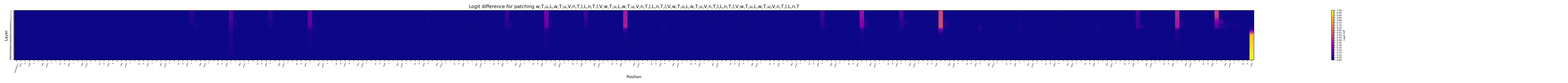}
    \caption{Different States Same Action: 30 Transitions}
    \label{fig:dfa-gpt-2s1a-30}
  \end{subfigure}

  \vspace{1em}

  \begin{subfigure}[b]{0.49\linewidth}
    \centering
    \includegraphics[width=\linewidth]{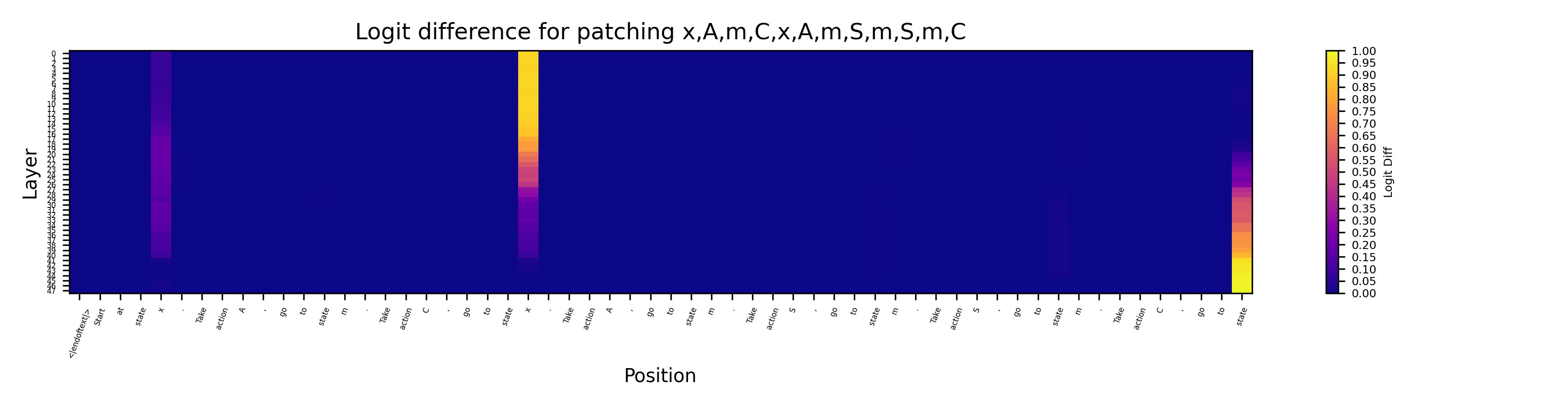}
    \caption{Long Sequences of Irrelevant Actions: 5 Transitions}
    \label{fig:dfa-gpt-noop-5}
  \end{subfigure}%
  \hfill
  \begin{subfigure}[b]{0.49\linewidth}
    \centering
    \includegraphics[width=\linewidth]{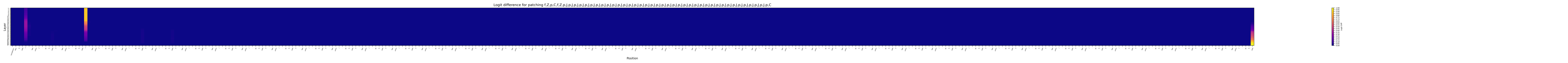}
    \caption{Long Sequences of Irrelevant Actions: 40 Transitions}
    \label{fig:dfa-gpy-noop-40}
  \end{subfigure}

  \caption{GPT2-XL Sequence Patching Heatmap}
  \label{fig:dfa-gpt-sequencepatching-grid}
\end{figure}

\subsection{Head Patching Heatmap: GPT2-XL}
\label{app:head-gpt2-xl}
\begin{figure}[H]
  \centering
  \begin{subfigure}[b]{0.49\linewidth}
    \centering
    \includegraphics[width=\linewidth]{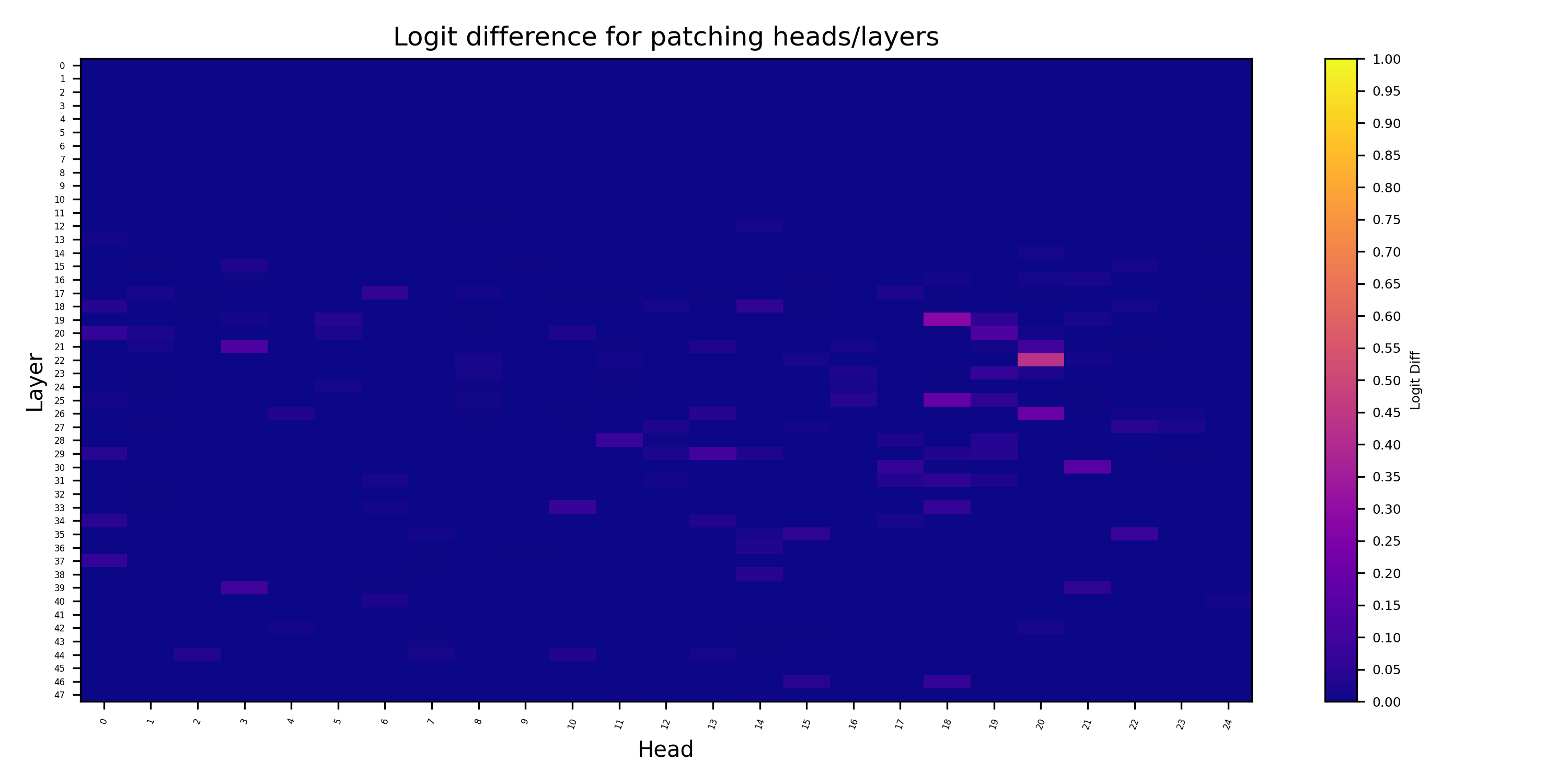}
    \caption{Different States Same Action}
    \label{fig:dfa-gpt-2s1a-head}
  \end{subfigure}%
  \hfill
  \begin{subfigure}[b]{0.49\linewidth}
    \centering
    \includegraphics[width=\linewidth]{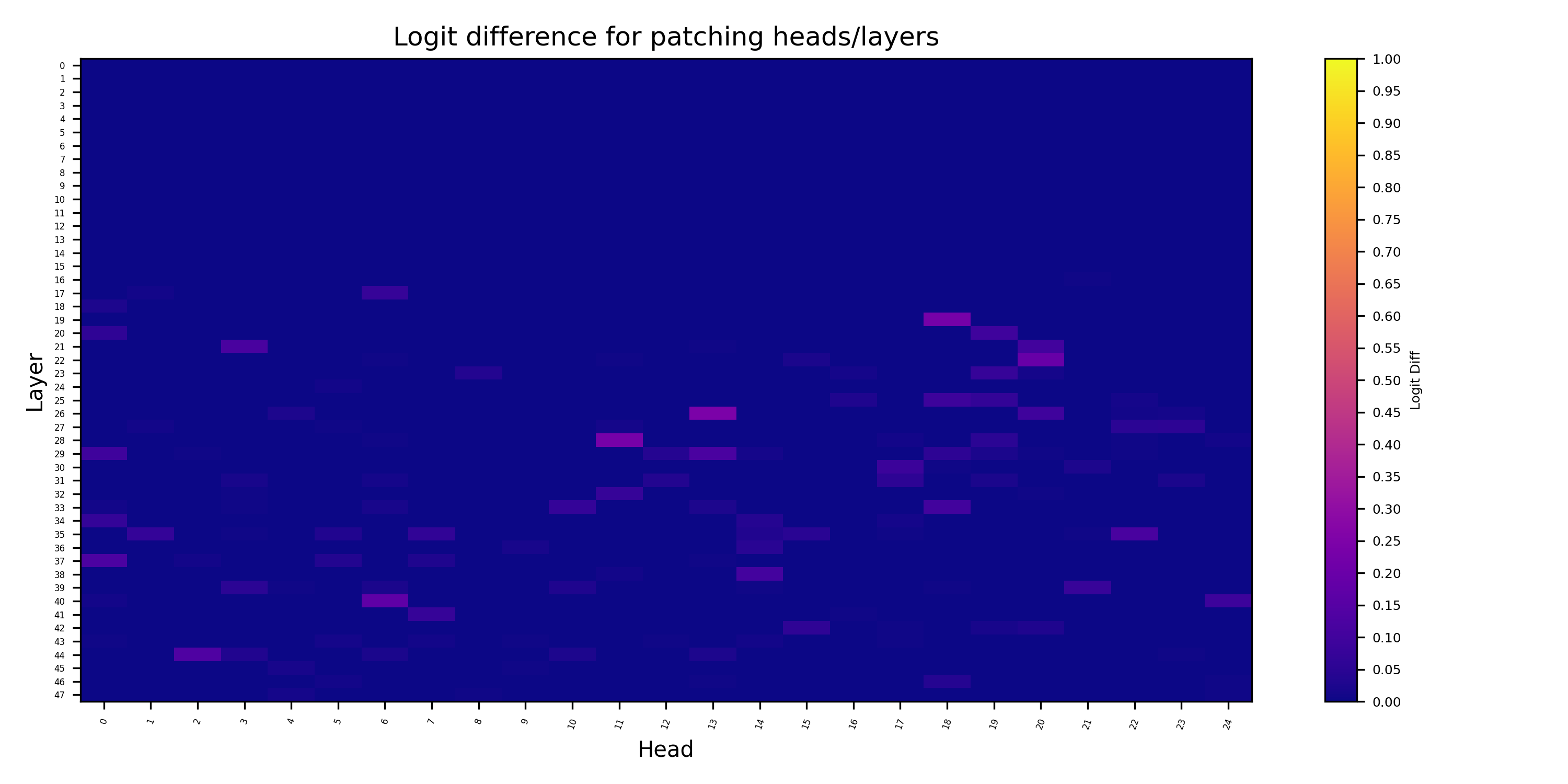}
    \caption{Long Sequences of Irrelevant Actions}
    \label{fig:dfa-gpt-noop-head}
  \end{subfigure}

  \caption{GPT2-XL Head Patching Heatmap}
  \label{fig:dfa-gpt-headpatching-grid}
\end{figure}

\subsection{Sequence Patching Heatmap: Pythia-1B}
\label{app:seq-pythia-1b}
\begin{figure}[H]
  \centering
  \begin{subfigure}[b]{0.49\linewidth}
    \centering
    \includegraphics[width=\linewidth]{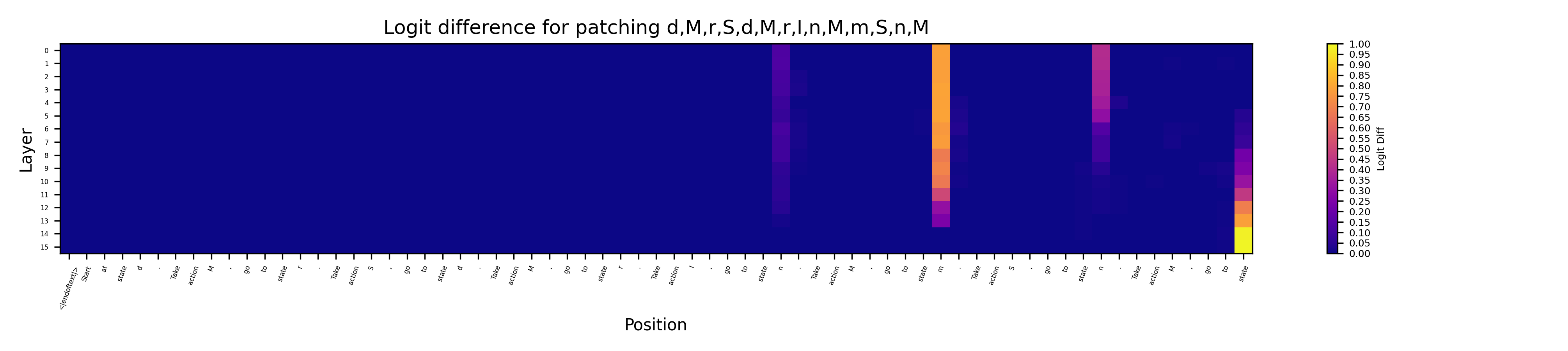}
    \caption{Different States Same Action: 6 Transitions}
    \label{fig:dfa-pythia-2s1a-6}
  \end{subfigure}%
  \hfill
  \begin{subfigure}[b]{0.49\linewidth}
    \centering
    \includegraphics[width=\linewidth]{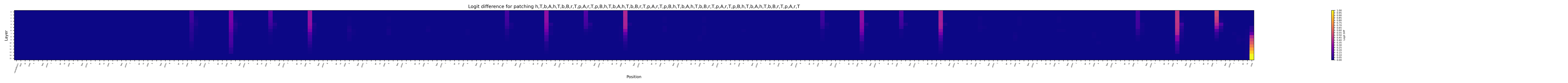}
    \caption{Different States Same Action: 30 Transitions}
    \label{fig:dfa-pythia-2s1a-30}
  \end{subfigure}

  \vspace{1em}

  \begin{subfigure}[b]{0.49\linewidth}
    \centering
    \includegraphics[width=\linewidth]{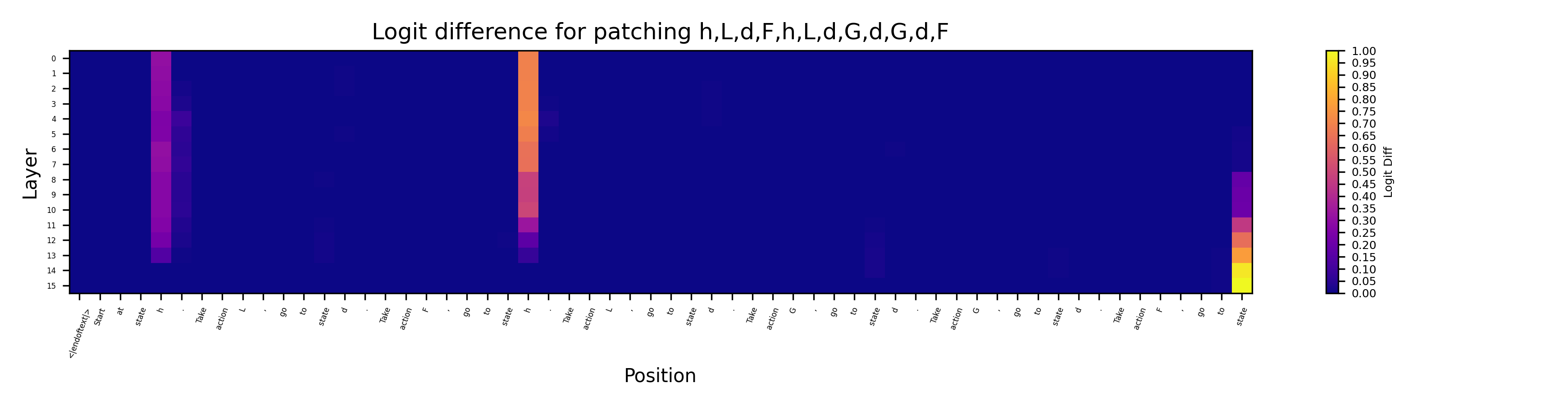}
    \caption{Long Sequences of Irrelevant Actions: 5 Transitions}
    \label{fig:dfa-pythia-noop-5}
  \end{subfigure}%
  \hfill
  \begin{subfigure}[b]{0.49\linewidth}
    \centering
    \includegraphics[width=\linewidth]{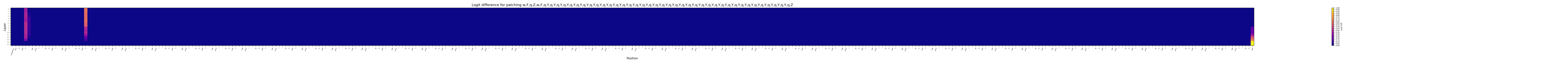}
    \caption{Long Sequences of Irrelevant Actions: 40 Transitions}
    \label{fig:dfa-pythia-noop-40}
  \end{subfigure}

  \caption{Pythia-1B Sequence Patching Heatmap}
  \label{fig:dfa-pythia-sequencepatching-grid}
\end{figure}

\subsection{Head Patching Heatmap: Pythia-1B}
\label{app:head-pythia-1b}
\begin{figure}[H]
  \centering
  \begin{subfigure}[b]{0.49\linewidth}
    \centering
    \includegraphics[width=\linewidth]{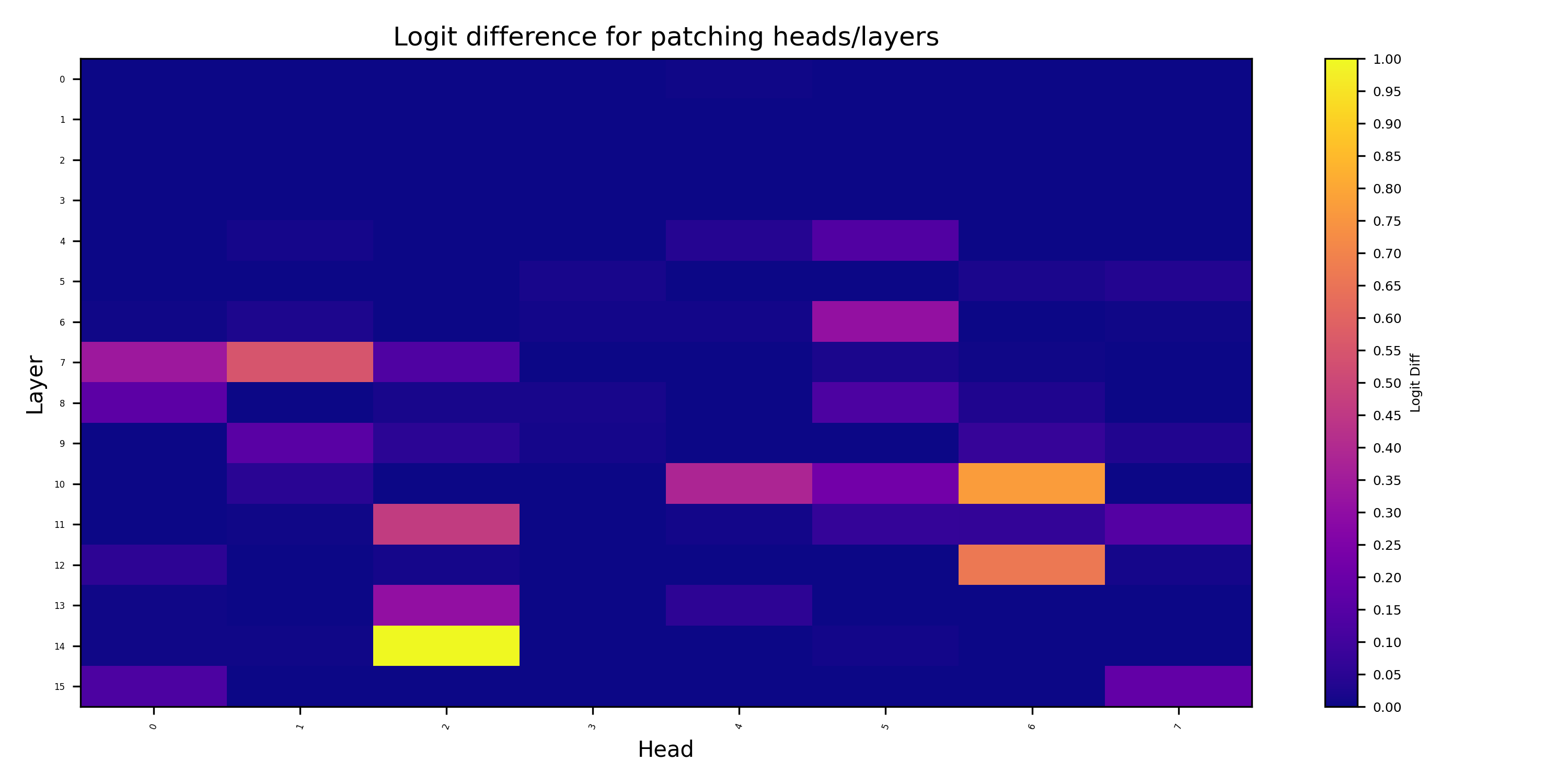}
    \caption{Different States Same Action}
    \label{fig:dfa-pythia-2s1a-head}
  \end{subfigure}%
  \hfill
  \begin{subfigure}[b]{0.49\linewidth}
    \centering
    \includegraphics[width=\linewidth]{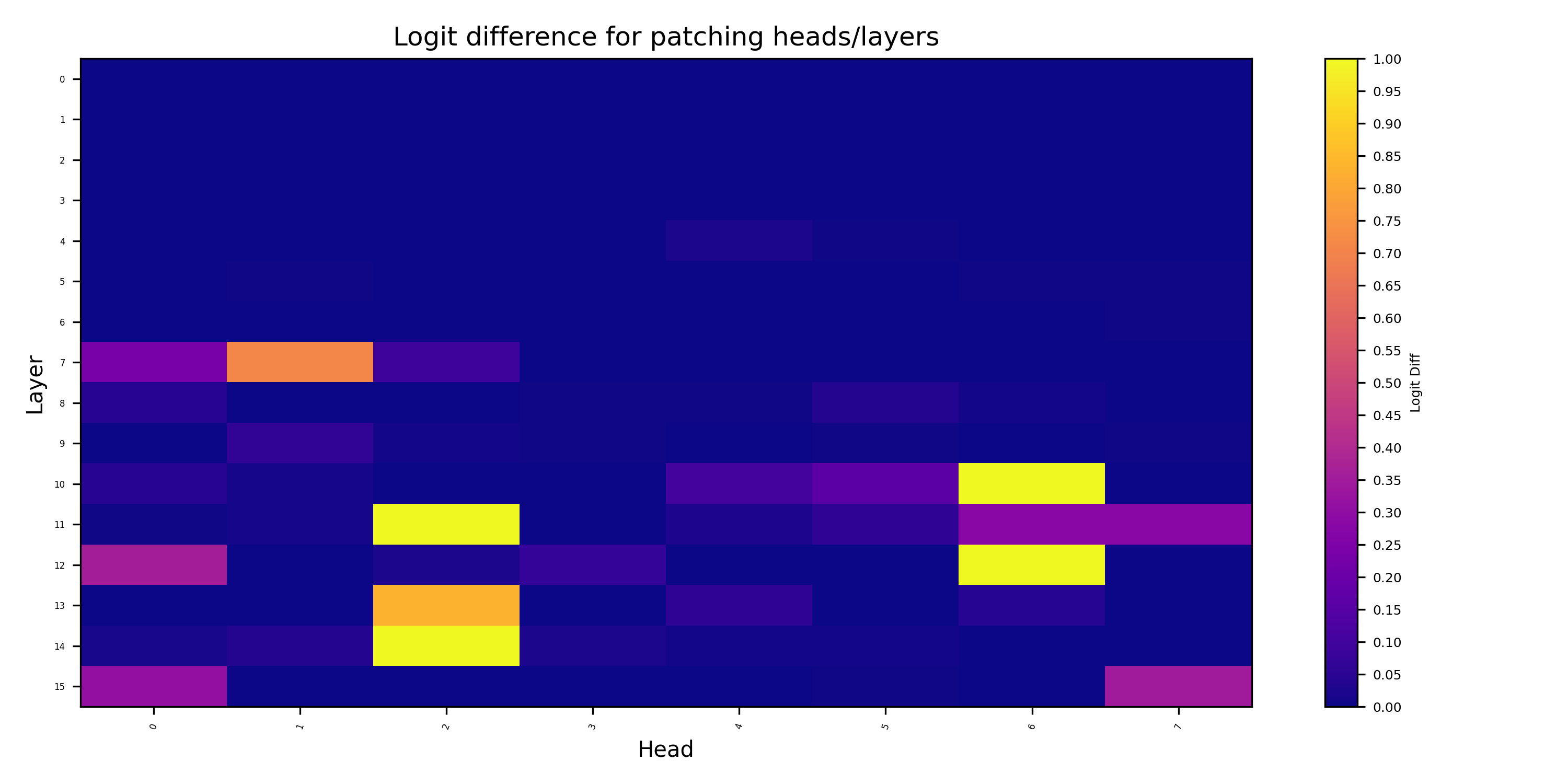}
    \caption{Pythia1B attention head patching for long Sequences of irrelevant actions}
    \label{fig:dfa-pythia-noop-head}
  \end{subfigure}
\caption{Pythia-1B Head Patching Heatmap}
  \label{fig:dfa-pythia-headpatching-grid}
\end{figure}

\subsection{Attention Pattern: DFA Different States Same Action, GPT-XL Head 20 Layer 22}
\label{app:a12}
\begin{figure}[H]
  \centering
    \includegraphics[width=0.5\linewidth]{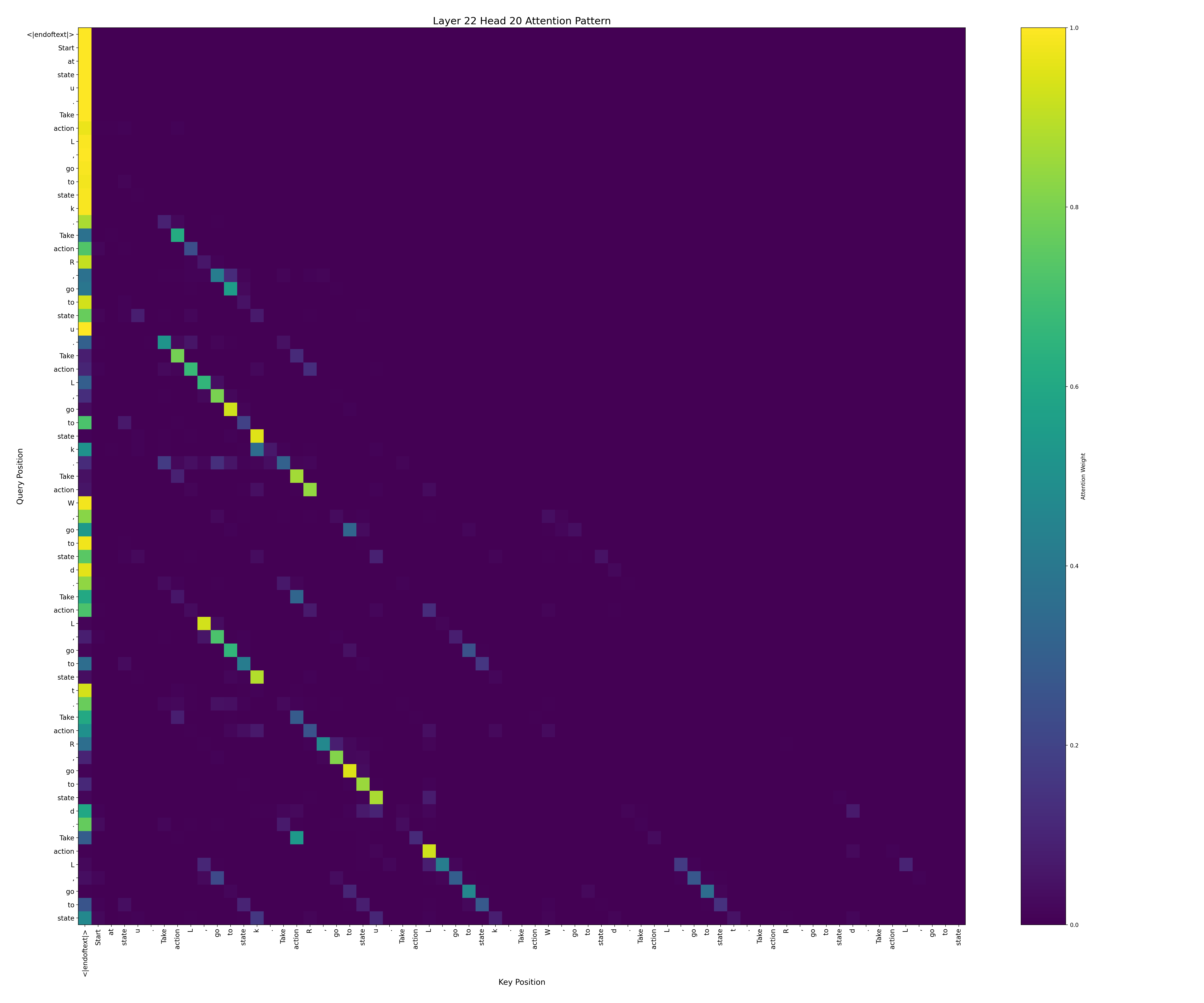}
    \caption{GPT-XL attention head patching for different state same action counterfactual}

\end{figure}

\subsection{Attention Pattern: DFA Long Sequences of Irrelevant Actions, GPT-XL Head 20 Layer 22}
\begin{figure}[H]
  \centering
    \includegraphics[width=0.5\linewidth]{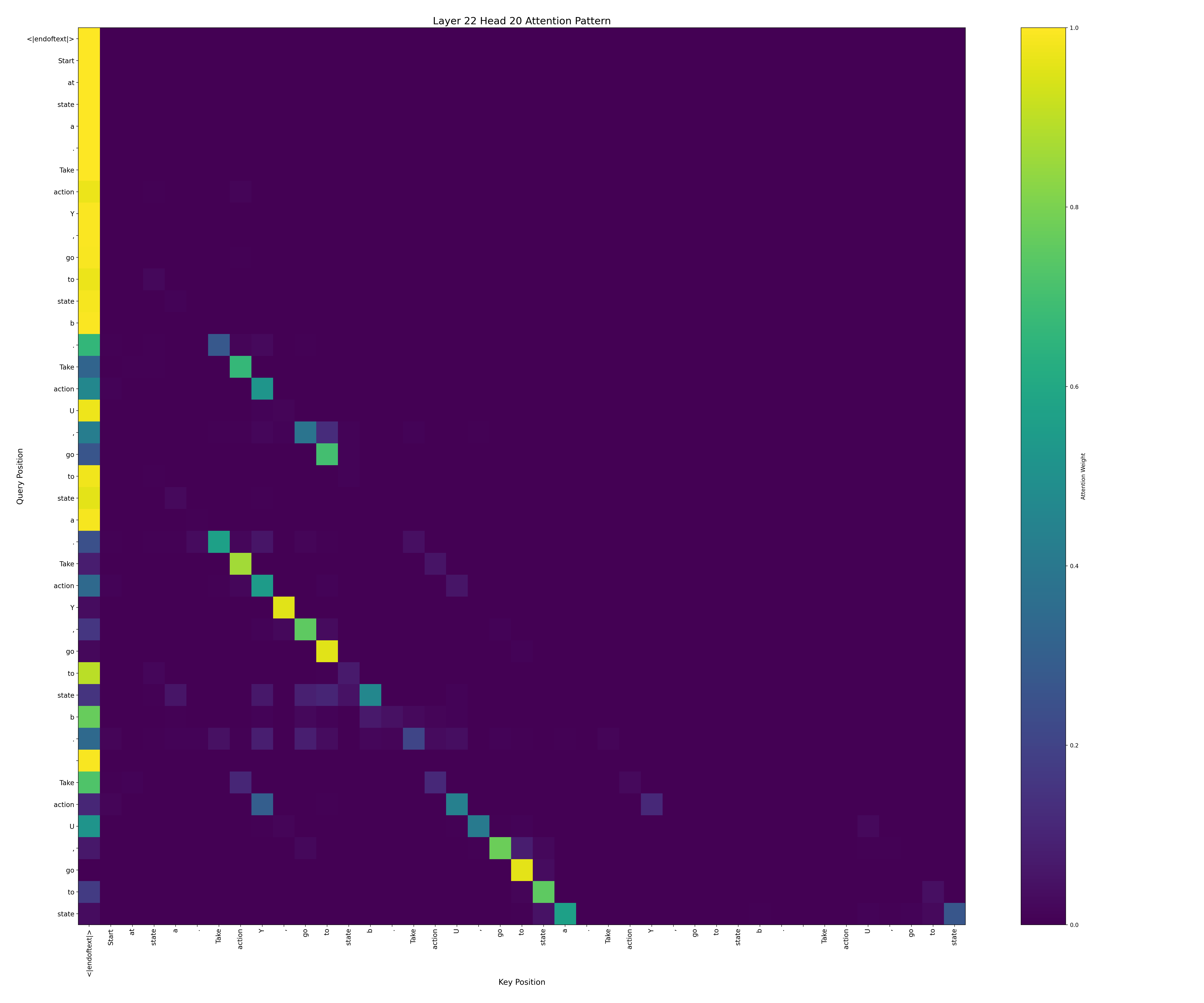}
    \caption{GPT-XL attention head patching for irrelevant action counterfactual}
\end{figure}

\subsection{Attention Pattern: DFA Different States Same Action, Pythia-1B}
\label{app:a13}
\begin{figure}[H]
  \centering
  \begin{subfigure}[b]{0.49\linewidth}
    \centering
    \includegraphics[width=\linewidth]{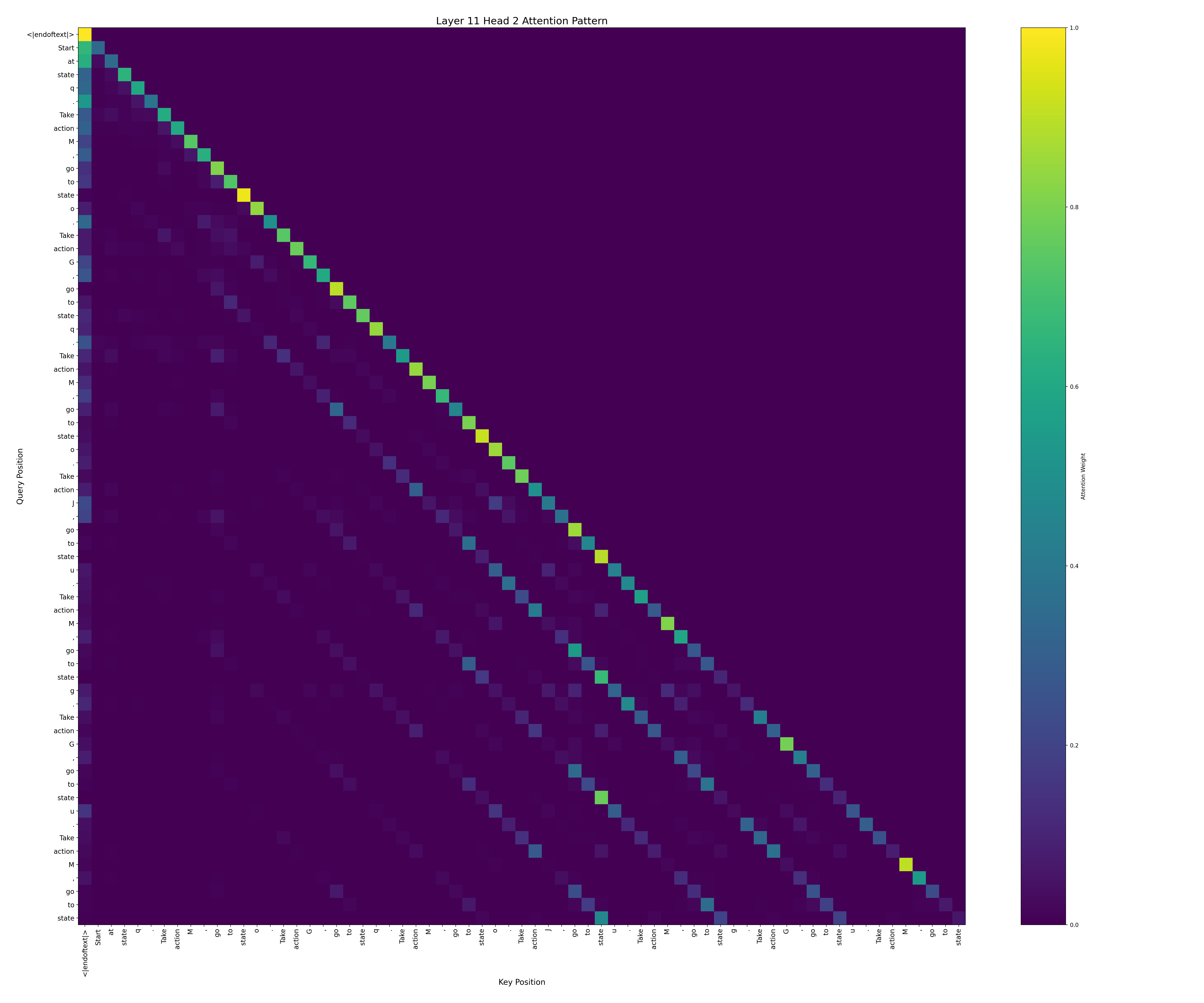}
    \caption{Head 2 Layer 11}
    \label{fig:dfa-pythia-attention-211}
  \end{subfigure}%
  \hfill
  \begin{subfigure}[b]{0.49\linewidth}
    \centering
    \includegraphics[width=\linewidth]{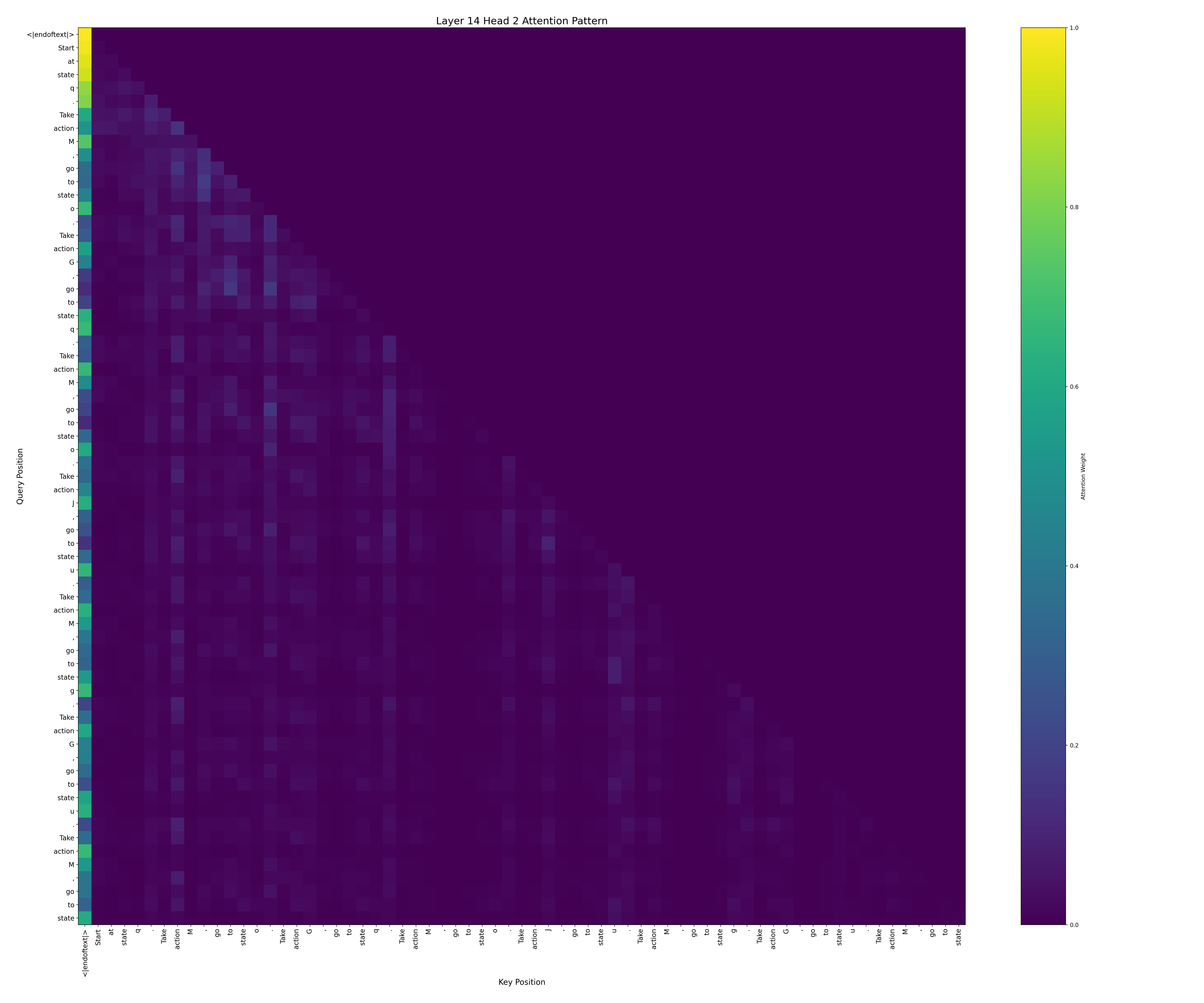}
    \caption{Head 2 Layer 14}
    \label{fig:dfa-pythia-attention-214}
  \end{subfigure}

  \vspace{1em}

  \begin{subfigure}[b]{0.49\linewidth}
    \centering
    \includegraphics[width=\linewidth]{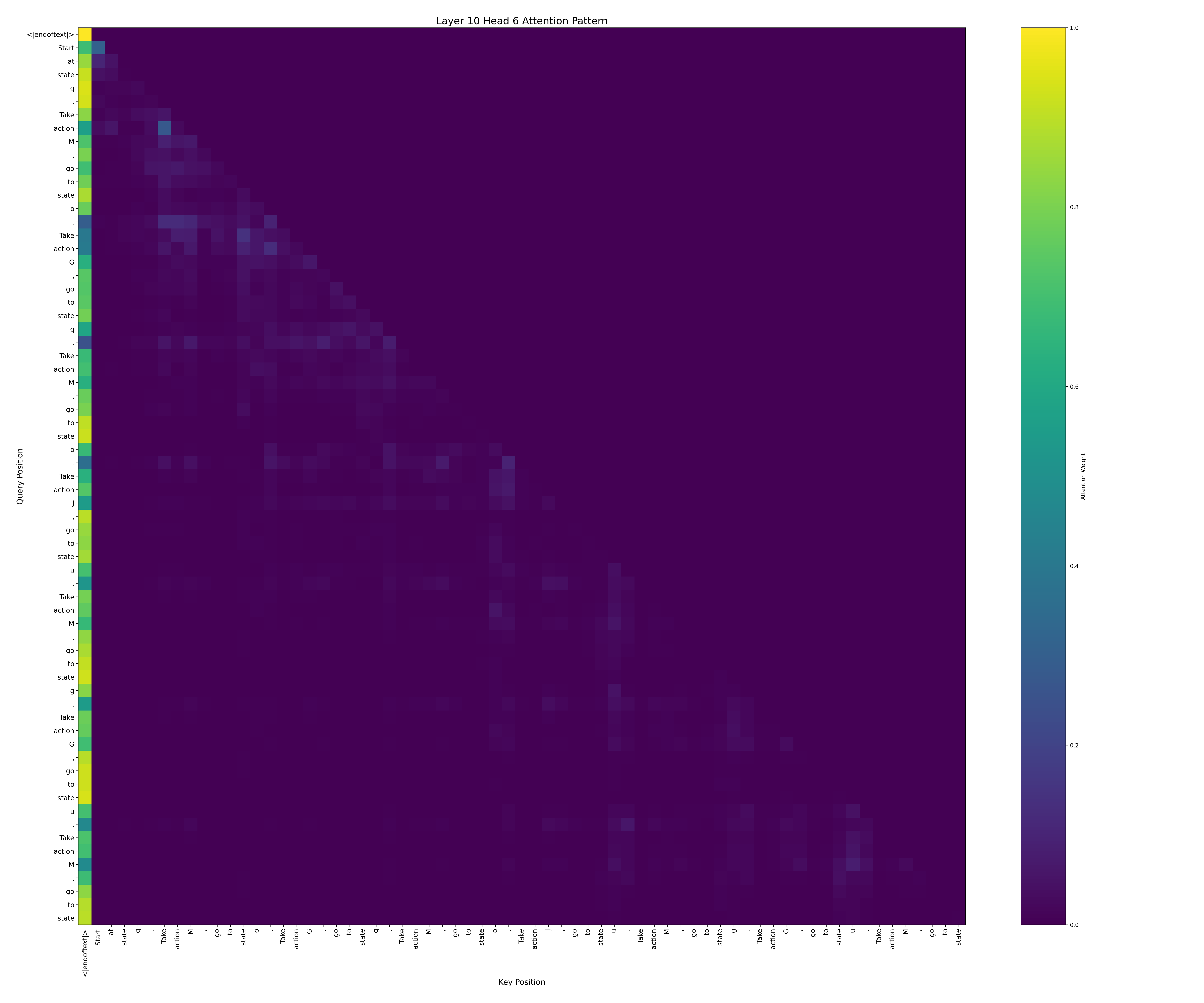}
    \caption{Head 6 Layer 10}
    \label{fig:dfa-pythia-attention-610}
  \end{subfigure}%
  \hfill
  \begin{subfigure}[b]{0.49\linewidth}
    \centering
    \includegraphics[width=\linewidth]{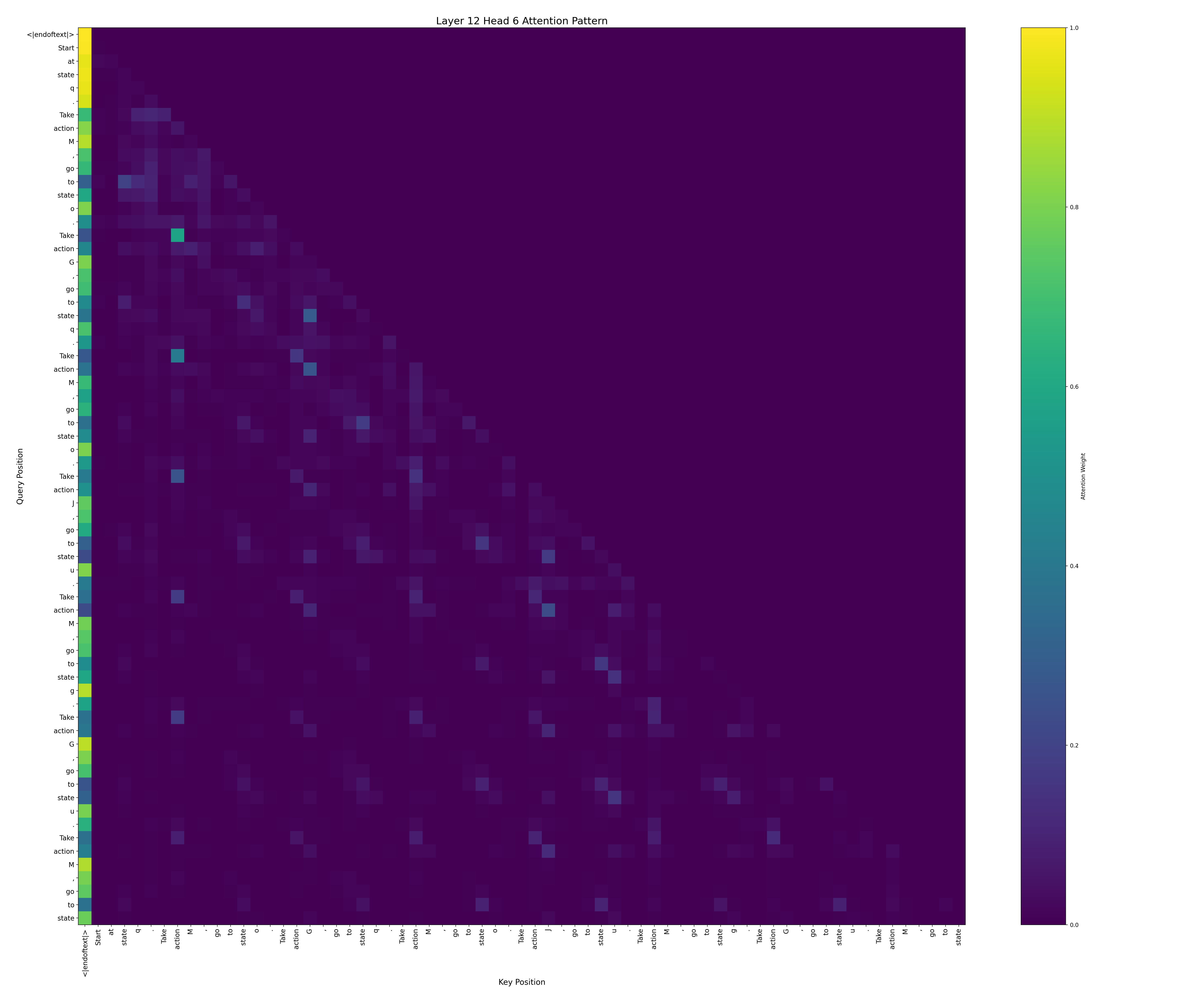}
    \caption{Head 6 Layer 12}
    \label{fig:dfa-pythia-attention-612}
  \end{subfigure}

  \caption{Pythia-1B attention head patterns for different state same action counterfactuals}
  \label{fig:dfa-pythia-attention-grid}
\end{figure}

\subsection{Attention Pattern: DFA Long Sequences 
of Irrelevant Actions, Pythia-1B}
\label{app:a14}
\begin{figure}[H]
  \centering
  \begin{subfigure}[b]{0.49\linewidth}
    \centering
    \includegraphics[width=\linewidth]{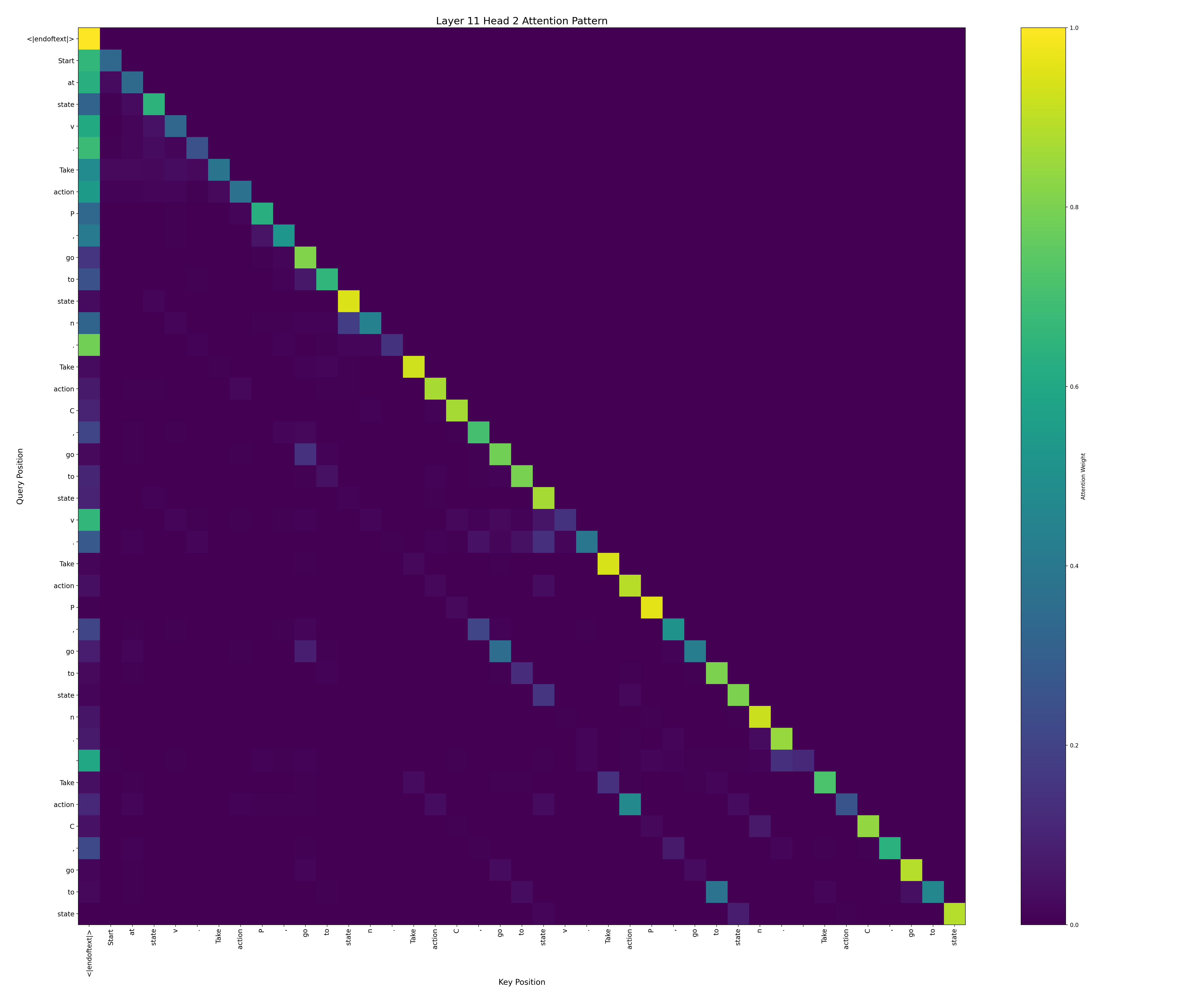}
    \caption{Head 2 Layer 11}
    \label{fig:dfa-pythia-attention-211}
  \end{subfigure}%
  \hfill
  \begin{subfigure}[b]{0.49\linewidth}
    \centering
    \includegraphics[width=\linewidth]{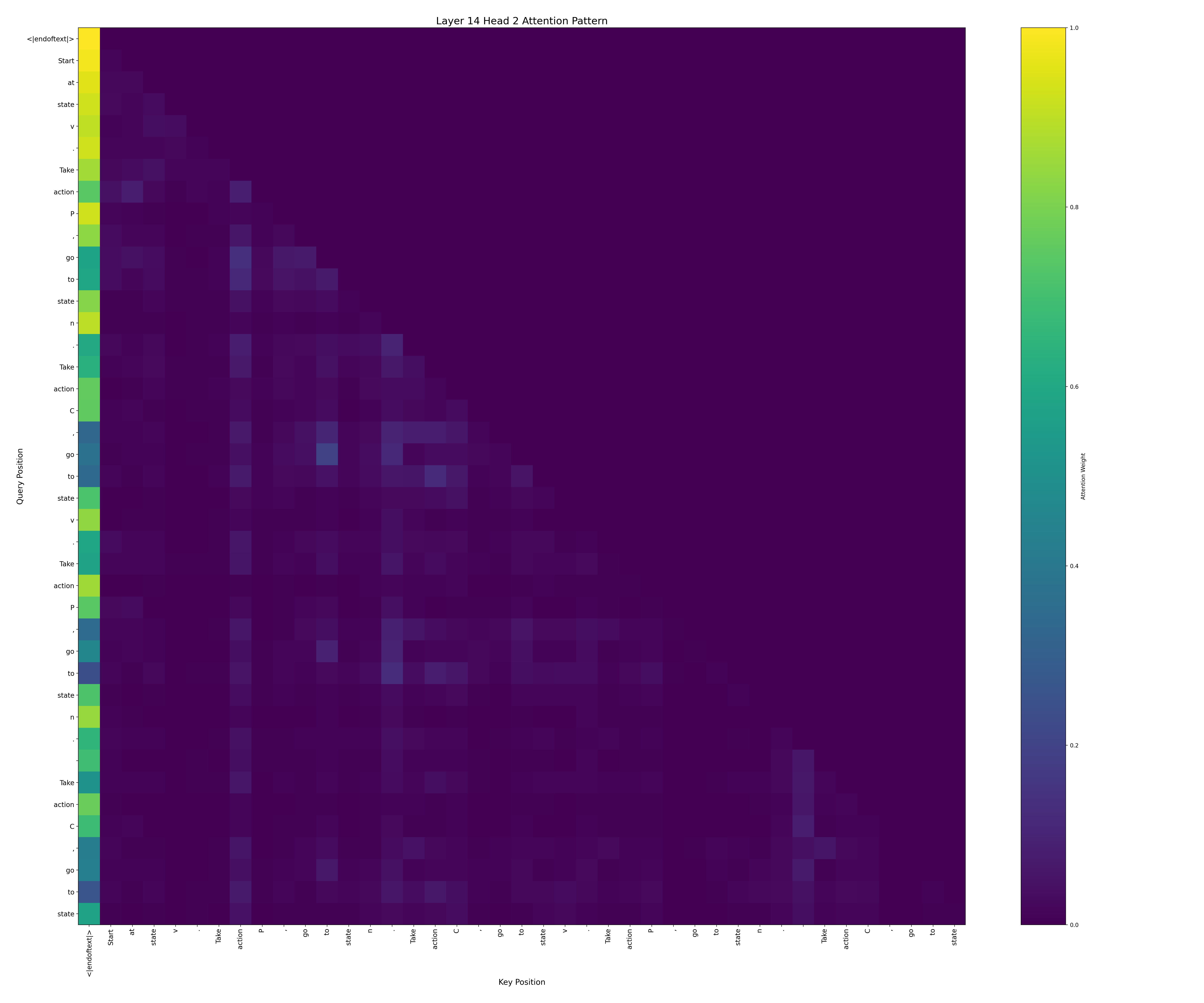}
    \caption{Head 2 Layer 14}
    \label{fig:dfa-pythia-attention-214}
  \end{subfigure}

  \vspace{1em}

  \begin{subfigure}[b]{0.49\linewidth}
    \centering
    \includegraphics[width=\linewidth]{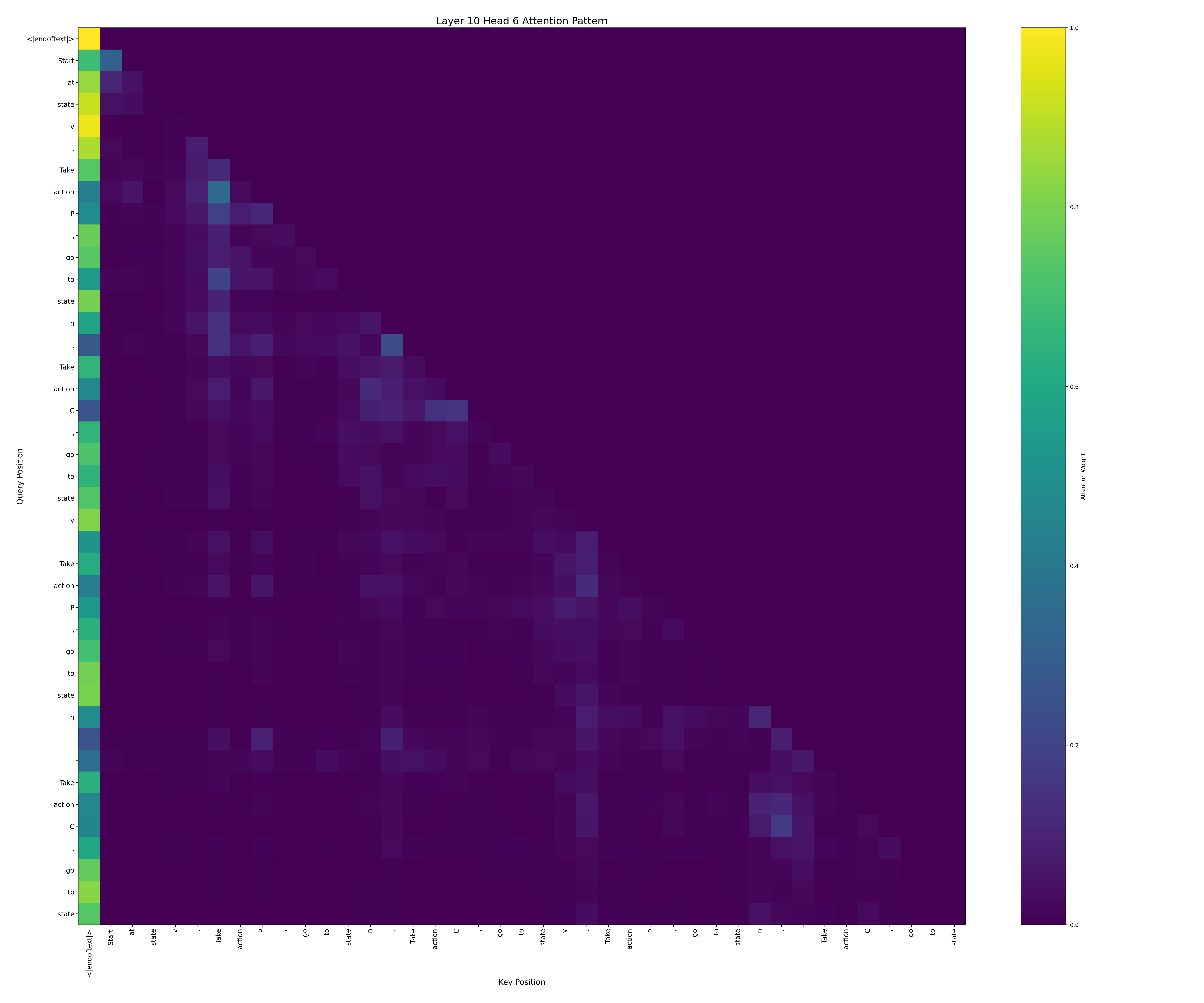}
    \caption{Head 6 Layer 10}
    \label{fig:dfa-pythia-attention-610}
  \end{subfigure}%
  \hfill
  \begin{subfigure}[b]{0.49\linewidth}
    \centering
    \includegraphics[width=\linewidth]{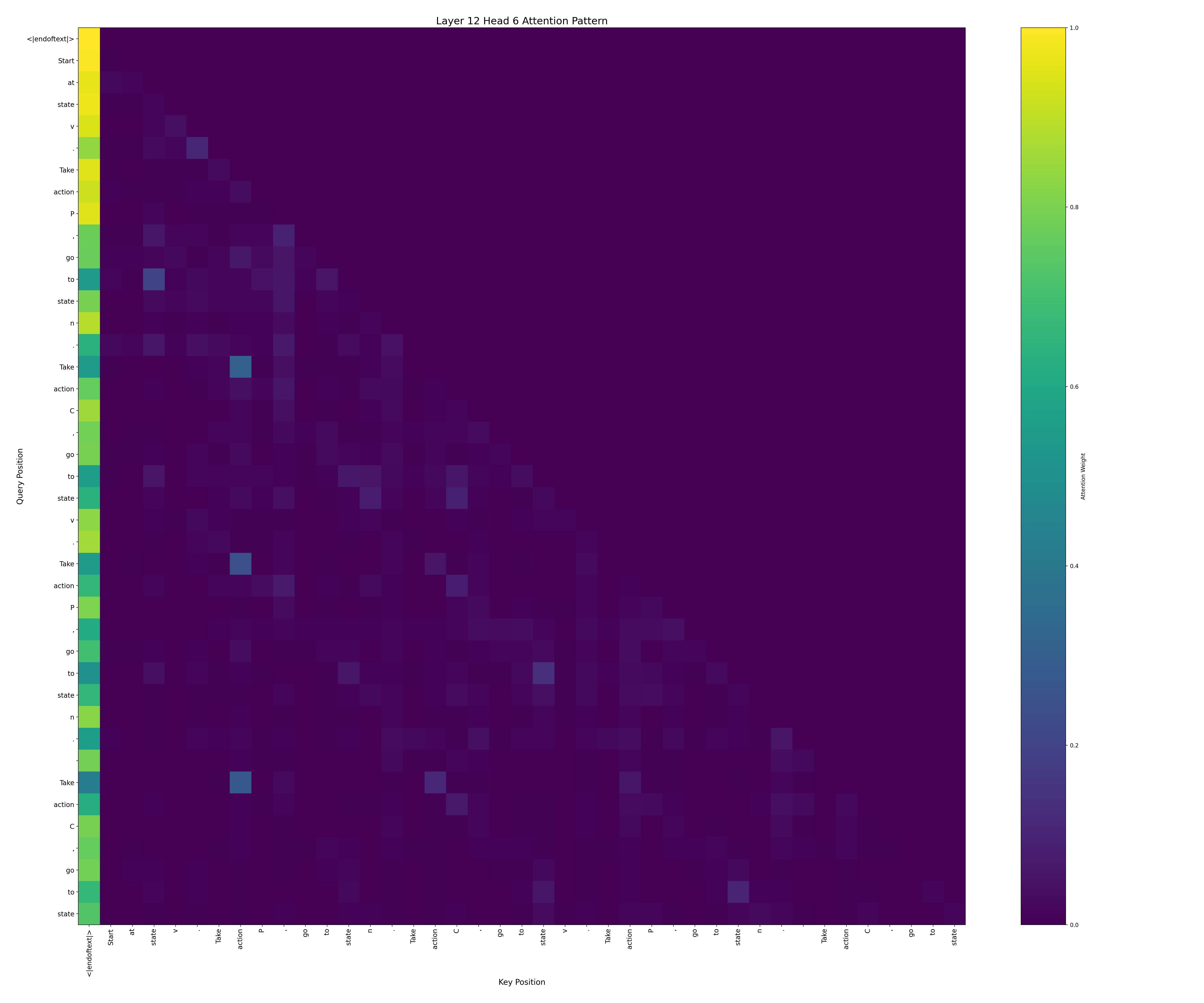}
    \caption{Head 6 Layer 12}
    \label{fig:dfa-pythia-attention-612}
  \end{subfigure}

\caption{Pythia-1B attention head patterns for sequences of irrelevant actions counterfactuals}
  \label{fig:dfa-pythia-attention-grid}
\end{figure}

\end{document}